\documentclass[10pt]{article} 
\usepackage[preprint]{tmlr}
\usepackage{hyperref}
\usepackage{url}
\usepackage{floatflt}
\usepackage{wrapfig}
\usepackage{graphicx} 
\usepackage{microtype}
\usepackage{algorithmic,enumitem}
\usepackage[ruled,vlined]{algorithm2e}
\usepackage{graphicx,array}
\usepackage{setspace}
\usepackage{subfigure,tabularx}
\usepackage{booktabs,caption} 
\usepackage{amssymb,amsmath,amsfonts,mathrsfs}
\usepackage{etoolbox,verbatim,color,adjustbox,booktabs}
\usepackage{pdfpages}
\usepackage{hyperref}
\usepackage{multirow}
\usepackage{makecell}
\usepackage{xcolor}  
\usepackage{ifthen}
\newtheorem{definition}{Definition}
\newtheorem{proposition}{Proposition}
\newtheorem{corollary}{Corollary}
\newtheorem{theorem}{Theorem}

\newcolumntype{S}{>{\footnotesize}c}

\DeclareMathOperator{\expect}{\mathbb{E}}

\newcommand{\mcalF}{\mathcal{F}} 
\newcommand{\mcalR}{\mathcal{R}} 
\newcommand{\mcalP}{\mathcal{P}}
\newcommand{\mcalU}{\mathcal{U}}
\newcommand{\mcalN}{\mathcal{N}} 
\newcommand{\mcalM}{\mathcal{M}} 
  
\newcommand{\mcalG}{\mathcal{G}}  
\newcommand{\mcalW}{\mathcal{W}}

\newcommand{\wtop}{w^\top}

\newcommand{\mbbR}{\mathbb{R}}   
\newcommand{\mbfR}{\mathbf{R}}   
\newcommand{\mfrakR}{\mathfrak{R}}

\newcommand{\zeroone}{0\text{-}1}

\newcommand{\Rnu}{\tilde{\mcalR}_{nu}}
\newcommand{\Ros}{\hat{\mcalR}^1_s}
\newcommand{\Rts}{\hat{\mcalR}^2_s}
\newcommand{\Rnn}{\hat\mcalR^-_n}
 \newcommand{\Rpp}{\hat{\mcalR}_p^+}
\newcommand{\Rup}{\hat\mcalR^+_u}
\newcommand{\Rnp}{\hat \mcalR^+_n}
\newcommand{\supgg}{\sup_{g\in \mcalG}}

\definecolor{brightpink}{rgb}{1.0, 0.0, 0.5}
\definecolor{blue}{rgb}{0.0, 0.0, 1.0}
\newcommand{\hien}[1]{{\color{black} #1}} 
 
\def\supm{1}

\title{Anomaly detection with semi-supervised 
 classification based on risk estimators}

 \author{\name  Le Thi Khanh Hien \email thikhanhhien.le@umons.ac.be \\
      \addr Department of Mathematics and Operational Research\\
      University of Mons, Belgium
      \AND
      \name Sukanya Patra \email sukanya.patra@umons.ac.be \\
      \addr Department of Computer Science,
  University of Mons, Belgium
      \AND
      \name Souhaib Ben Taieb \email souhaib.bentaieb@umons.ac.be\\
      \addr Department of Computer Science,
  University of Mons, Belgium}
      
\def\month{MM}  
\def\year{YYYY} 
\def\openreview{\url{https://openreview.net/forum?id=XXXX}} 

\begin{document}
\maketitle
\begin{abstract}
A significant limitation of one-class classification anomaly detection methods is their reliance on the assumption that unlabeled training data only contains normal instances. To overcome this impractical assumption, we propose two novel classification-based anomaly detection methods. Firstly, we introduce a semi-supervised shallow anomaly detection method based on an unbiased risk estimator. Secondly, we present a semi-supervised deep anomaly detection method utilizing a nonnegative (biased) risk estimator. We establish estimation error bounds and excess risk bounds for both risk minimizers. Additionally, we propose techniques to select appropriate regularization parameters that ensure the nonnegativity of the empirical risk in the shallow model under specific loss functions. Our extensive experiments provide strong evidence of the effectiveness of the risk-based anomaly detection methods.
\end{abstract}

\section{Introduction} 
Anomaly Detection (AD) can be defined as the task of identifying instances that deviates significantly from the majority of the data instances, see e.g.,    \citep{Chandola2009AnomalyDetection,Pang2020DeepReview,Ruff2021_review} for comprehensive surveys on AD. One important approach for AD is one-class classification \cite{khan_madden_2014,Tax1999SupportDescription}. It can be viewed as a specialized binary classification problem aimed at learning a model that distinguishes between positive (normal) and negative (anomalous) classes. This approach assumes that the unlabeled dataset primarily consists of data from the normal class. By utilizing a sufficient amount of normal data, one-class classification AD (OC-AD) methods identify a decision boundary that encompasses all the normal points. For example, the decision boundaries of \emph{shallow} OC-AD methods include a hyperplane with maximum margin \cite{Scholkopf2001}, a compact spherical boundary \cite{Tax1999SupportDescription, Tax2004_SVDD}, an elliptical boundary \citep{Rousseeuw1999AEstimator,Rousseeuw1985MultivariatePoint}, a pair of subspaces \cite{WangICCV2019}, or even a collection of multiple spheres \cite{Gornitz2018}. To enhance their applicability in high-dimensional settings, these shallow methods have been extended into \emph{deep} methods \cite{Erfani2016High-dimensionalLearning,Ruff2018DeepClassification}.

Unsupervised learning, where only unlabeled data is available, represents the most common setting in AD. Unsupervised AD methods typically assume that the training data consists solely of normal instances \cite{Hodge2004,PIMENTEL2014,Zimek2012}. However, in real-world scenarios, labeled samples may be available alongside the unlabeled dataset, leading to the development of semi-supervised AD methods, including semi-supervised OC-AD methods \cite{Gornitz2009,Mari2010,Ruff2020Deep}. It is important to note that unsupervised/semi-supervised shallow/deep one-class anomaly detection  methods do not explicitly handle mixed unlabeled data. This is because they typically assume that there are no anomalous instances present in the unlabeled dataset, which is impractical in real-world scenarios. Classification methods that handle mixed unlabeled data have been extensively studied in the field of learning with positive and unlabeled examples (LPUE or PU learning). In this context, we have access to information on positive and unlabeled data, but negative data is unavailable. PU learning methods have also been utilized as semi-supervised AD methods \cite{Bekker2020,Blanchard2010,Chandola2009AnomalyDetection,JU2020}. It is widely recognized that incorporating labeled anomalies, even if only a few instances, can greatly enhance the AD performance \cite{GornitzKRB13,Kiran2018}. Semi-supervised AD methods that consider the availability of negative data have demonstrated highly promising AD performance \cite{han2022adbench,Ruff2021_review,Ruff2020Deep}.

To overcome the impractical assumption of OC-AD methods, we adopt the key concept of risk-based PU learning methods \cite{Plessis_NIPS2014,Plessis_ICML2015,Kiryo_NIPS2017,Sakai_ICML2017}. These methods propose empirical estimators for the risk associated with the learning problem. In order to improve anomaly detection performance, we focus on the semi-supervised setting where a negative dataset is also available.

\textbf{Contributions} \quad Our main contributions are summarized as follows.
\begin{itemize}[leftmargin=0.5cm]
    \item  Considering AD as a semi-supervised binary classification problem, where we have access to a positive dataset, a negative dataset, and an unlabeled dataset that may contain anomalous examples, we introduce two risk-based AD methods. These methods include a shallow AD approach developed using an unbiased risk estimator and a deep AD method based on a nonnegative risk estimator.

    \item We  develop methods to select suitable regularization that ensures the nonnegativity of the empirical risk in the proposed shallow AD method. This is crucial as negative empirical risk can lead to significant overfitting issues \cite{Kiryo_NIPS2017}.

    \item We additionally establish estimation error bounds and excess risk bounds for the two risk minimizers, building upon the theoretical findings presented in \cite{Kiryo_NIPS2017,Niu2016}.

    \item  We conduct extensive experiments on benchmark AD datasets obtained from \emph{Adbench} \cite{han2022adbench} to compare the performance of our proposed risk-based AD (rAD) methods against various baseline methods.

\end{itemize}

\textbf{Organization} \quad In Section \ref{sec:review}, we provide a brief background on risk estimators. We then introduce the two risk estimators that form the basis of our risk-based AD methods in Section \ref{sec:rAD-estimators}. Additionally, we present a theoretical analysis in Section \ref{sec:risk_bound}, discuss related work in Section \ref{sec:related}, present experimental results in Section \ref{sec:experiments}, highlight limitations in Section \ref{sec:limitation}, and conclude the paper in Section \ref{sec:conclusion}. All proofs and additional experiments can be found in the supplementary material.  

\section{Background on risk estimators}
\label{sec:review}
Let $x$ and $y \in \{+1, -1\}$ be random variables 
with joint density $p(x, y)$. The class-conditional densities are $p_p(x)=P(x|y=+1)$ and $p_n(x)=P(x|y=-1)$. Let $\pi_p=p(y=+1)$ and $\pi_n=p(y=-1)$ be the class-prior probabilities for the positive and negative classes. We have $ \pi_p+\pi_n=1$. Suppose the positive $(\mcalP)$, negative $(\mcalN)$ and unlabeled $(\mcalU)$ data are sampled independently as 
$(\mcalP) = \{x_i^p\}_{i=1}^{n_p} \sim p_p(x), (\mcalN) = \{x_i^n\}_{i=1}^{n_n} \sim p_n(x),  
 (\mcalU) = \{x_i^u\}_{i=1}^{n_u} \sim p(x),$ 
 where 
 \begin{equation}
 \label{eq:px}
 p(x)=\pi_p p_p(x) + \pi_n p_n(x).
 \end{equation}
 Given $(\mcalP)$,  $(\mcalN)$ and $(\mcalU)$, let us consider a binary classification problem from $x$ to $y$. Suppose $g: \mathbb R^d \to \mathbb R$ is a decision function that needs to be trained from  $(\mcalP)$,  $(\mcalN)$ and $(\mcalU)$, and  $\ell : \mathbb R \times \{+1,-1\} \to  \mathbb R$ is a loss function that imposes a cost $ \ell(t,y)$ if the predicted output is $t$ and the expected output is $y$. Under loss $\ell$, let us denote
\[ 
\begin{split}
&\mathcal R_p^+(g)=\expect_{x \sim p_p(x)}[\ell(g(x),+1)], \mathcal R_n^+(g)=\expect_{x \sim p_n(x)}[\ell(g(x),+1)], {\mcalR}_{u}^+(g)=\expect_{x \sim p(x)}[\ell (g(x),+1)],\\
&\mathcal R_p^-(g)=\expect_{x \sim p_p(x)}[\ell(g(x),-1)], \mathcal R_n^-(g)=\expect_{x \sim p_n(x)}[\ell(g(x),-1)],
 \mcalR_{u}^-(g)=\expect_{x \sim p(x)}[\ell (g(x),-1)]. 
\end{split}
\]
 \hien{Given $\ell$ and assuming that  $\pi_p$ is known (in practice, $\pi_p$ can be effectively estimated from $(\mcalP)$,  $(\mcalN)$ and $(\mcalU)$ \cite{Plessis2013,Saerens2002})}, our goal is to find $g$ that minimizes the risk of $g$, which is defined by 
\begin{equation}
\label{def:risk}
\mcalR(g):=\expect_{(x,y)\sim p(x,y)} [\ell (g(x),y)] = \pi_p \mcalR^+_p(g) + \pi_n \mcalR^-_n(g).
\end{equation}
In ordinary classification, the optimal classifier minimizes the expected
misclassification rate that corresponds to using zero-one loss in \eqref{def:risk}, $\ell_{\zeroone}(t,y)=0$ if $ty>0$ and $\ell_{\zeroone}(t,y)=1$ otherwise. We denote $I(g)=\expect_{(x,y)\sim p(x,y)} [\ell_{\zeroone} (g(x),y)]$.

\textbf{PN risk estimator} \quad In supervised learning when we have fully labeled data,  $\mcalR(g)$ can be approximated by a PN risk estimator 
$ \hat{\mcalR}_{pn}(g)=\pi_p\hat{\mcalR}^+_p(g) + \pi_n\hat{\mcalR}_n^-(g),$ where  
\begin{equation}
\label{define_mcalR}
 \hat{\mcalR}^+_p(g):= \frac{1}{n_p} \sum_{i=1}^{n_p}\ell(g(x_i^p),+1), \quad \hat{\mcalR}_n^-(g):=\frac{1}{n_n} \sum_{i=1}^{n_n}\ell(g(x_i^n),-1). 
\end{equation}
\textbf{PU risk estimator}\quad In PU 
 learning when $(\mathcal N)$ is unavailable, \cite{Plessis_NIPS2014,Plessis_ICML2015,Kiryo_NIPS2017} propose methods to approximate $\mcalR(g)$ from $(\mcalP)$ and $(\mcalU)$. 
From  \eqref{eq:px} we have 
$\pi_n \mcalR^-_n(g)=\mcalR^-_u(g) -\pi_p  \mcalR^-_p(g),
$
which implies that 
\begin{equation}
\label{risk_1}
\mcalR(g)=\pi_p( \mcalR_p^+ - \mcalR_p^-)+ \mcalR_u^-(g).
\end{equation}
When $\ell$ satisfies the symmetric condition $\ell(t,+1)+\ell(t,-1)=1$ 
then we have $ \mcalR(g)=2\pi_p {\mcalR}_p^+(g) - \pi_p + {\mcalR}_u^-(g)$, 
which can be approximated by 
\begin{equation}
\label{eq:nonconvex_Rpu}
\hat{\mcalR}^1_{pu}(g)=2\pi_p \Rpp(g) - \pi_p + \hat{\mcalR}_u^-(g),
\end{equation}
where $\hat{\mcalR}^+_p(g)$ is defined in \eqref{define_mcalR} and $\hat{\mcalR}_u^-(g)= \frac{1}{n_u} \sum_{i=1}^{n_u}\ell(g(x_i^u),-1)$, see \cite{Plessis_NIPS2014}. When $\ell$ satisfies the linear-odd condition $\ell(t,+1)-\ell(t,-1)=-t$ 
then  $\mcalR(g)$ can be approximated by 
\begin{equation}
\label{eq:convex_Rpu}
\hat{\mcalR}^2_{pu}(g)=-\pi_p \frac{1}{n_p} \sum_{i=1}^{n_p} g(x_i^p) + \hat{\mcalR}_u^-(g),
\end{equation}  
see \cite{Plessis_ICML2015}. 
The authors in \cite{Kiryo_NIPS2017} propose a \emph{non-negative} PU risk estimator 
\begin{equation}
    \label{nnPU-risk}
\hat{\mcalR}^3_{pu}(g)= \pi_p \Rpp(g) + \max\{0, \hat{\mcalR}_u^-(g)-\pi_p \hat{\mcalR}_p^-(g)\},\end{equation} 
where $\hat{\mcalR}_p^-(g)=\frac{1}{n_p} \sum_{i=1}^{n_p}\ell(g(x_i^p),-1)$. 
Note that $\hat{\mcalR}^3_{pu}(g)$ is a biased estimator. 

\textbf{NU risk estimator}\quad Similarly, considering NU learning when $(\mathcal P)$ is unavailable, see \cite{Sakai_ICML2017}, NU risk estimators can be formulated by combining the equation 
$\pi_p \mcalR^+_p(g)=\mcalR^+_u(g) -\pi_n  \mcalR^+_n(g)$ (which is derived from  \eqref{eq:px} ) and  \eqref{def:risk} to obtain  
\begin{equation}
\label{risk_2}
\mcalR(g)=  - \pi_n( \mcalR_n^+ - \mcalR_n^-)+\mcalR^+_u(g).
\end{equation} 
\hien{With a loss satisfying the symmetric condition}, we have a nonconvex NU risk estimator 
\begin{equation}
\label{eq:nonconvex_Rnu}
\hat{\mcalR}^1_{nu}(g)=2\pi_n \hat{\mcalR}_n^-(g) - \pi_n + \hat{\mcalR}_u^+(g),
\end{equation}
where $\hat{\mcalR}_n^-(g) $ is defined in \eqref{define_mcalR} and $\hat{\mcalR}_u^+(g)=\frac{1}{n_u}\sum_{i=1}^{n_u} \ell(g(x_i^u),+1)$. 
\hien{And with a loss satisfying} the linear-odd condition, we get a convex NU risk estimator 
 \begin{equation}
\label{eq:convex_Rnu}
\hat{\mcalR}^2_{nu}(g)=\pi_n \frac{1}{n_n} \sum_{i=1}^{n_n} g(x_i^n) + \hat{\mcalR}_u^+(g).
\end{equation} 
Finally, \cite{Sakai_ICML2017} proposes to use a linear combination between the PN, the NU, and the PU risk of\cite{Plessis_NIPS2014,Plessis_ICML2015}.

\section{The proposed semi-supervised anomaly detection methods}
\label{sec:rAD-estimators}  
In the previous section, we presented risk estimators for the PU learning problem where $(\mcalN)$ is unavailable. Let us consider the setting where we have access to $(\mcalP)$, $(\mcalN)$ as well as $(\mcalU)$. We perceive semi-supervised AD as a binary classification problem from $x$ to $y\in \{+1,-1\}$, where $+1$ represents the normal class and $-1$ represents the anomalous class. Our goal is to propose risk estimators for the risk in \eqref{def:risk}. Specifically, we propose two risk estimators for semi-supervised AD that lead to  two risk-based AD methods.


If we take a convex combination of \eqref{def:risk} and \eqref{risk_2}, we obtain
\begin{align}
\mcalR(g)&=a( - \pi_n( \mcalR_n^+ - \mcalR_n^-) + \mcalR^+_u(g)) + (1-a) ( \pi_p \mcalR^+_p(g) + \pi_n \mcalR^-_n(g)) \nonumber\\
&=a \mcalR^+_u(g) + (1-a) \pi_p \mcalR^+_p(g) + \pi_n \mcalR_n^-(g) - a \pi_n \mcalR_n^+, \label{eq:conv}  
\end{align}
where $a\in (0,1)$. 

The empirical version of \eqref{eq:conv} yields the following linear combination of PN and NU risk estimators:
\begin{equation}
\label{risk_sem_1}
\Rts(g)= a \Rup(g) + (1-a) \pi_p \hat \mcalR^+_p(g) + \pi_n \hat \mcalR_n^-(g) - a \pi_n \hat\mcalR_n^+(g).
\end{equation}

While $\Rts(g)$ was also considered in \cite{Sakai_ICML2017}, they only focused on the set of linear classifiers with two specific losses -- the (scaled) ramp loss and the truncated (scaled) squared loss (see \cite[Section 4.1]{Sakai_ICML2017}). We consider a more general setting for $\Rts$ and also propose methods to choose appropriate regularization for $\Rts$ to avoid negative empirical risks. In fact, $\Rts$ may take negative values when $\ell$ is unbounded due to the negative term $-a\pi_n\hat\mcalR_n^+(g)$. Theorem \ref{thrm:nonnegative} summarizes the conditions that guarantee a nonnegative objective.



Inspired by $\hat{\mcalR}^3_{pu}(g)$ in \eqref{nnPU-risk}, we also propose the following nonnegative risk estimator: 
\begin{equation}
\label{risk_sem_2}
\Ros(g)= \pi_n \Rnn(g) + (1-a) \pi_p \hat\mcalR^+_p(g)  + a \max\{ 0, \Rup(g)-\pi_n \Rnp(g)  \},
\end{equation}
where the $\max$ term is introduced since $ \mcalR^+_u(g)-\pi_n \mcalR^+_n(g) =\pi_p \mcalR_p^+$ must be nonnegative. Note that $\hat{\mcalR}^3_{pu}(g)$ was designed for the  PU learning problem while we propose $\Ros(g)$ for the AD problem which often assumes anomalies are  rare. In other words, we put more emphasis on $ \Rup(g)$ rather than $\hat\mcalR^-_u(g)$.



 \hien{In Section \ref{sec:risk_bound}, we will establish the theoretical estimation error bounds and excess risk bounds for the minimizers of both  $ \min_{g\in\mathcal G} \Ros(g)$ and $ \min_{g\in\mathcal G} \Rts(g)$, where $\mcalG$ is some class function. We now present the practical optimization problems involved when using $\Ros(g)$ and $\Rts(g)$.}
 
\hien{\textbf{Optimization problems} \quad 
 Suppose $g$ is parameterized by $w$, which needs to be learned from $(\mcalP)$, $(\mcalN)$ and $(\mcalU)$.}
When  $\Ros$ in \eqref{risk_sem_2} is used,  the corresponding optimization problem for AD  is  
\begin{equation}
\label{opt_1_g}
\begin{split}
&\min_{w} \Big\{ \frac{ \pi_n}{n_n} \sum_{i=1}^{n_n}\ell(g(x_i^n),-1) + \frac{ (1-a) \pi_p}{n_p} \sum_{i=1}^{n_p}\ell(g(x_i^p),+1) \\
&\qquad\qquad + a \max \big\{ 0, \frac{1}{n_u} \sum_{i=1}^{n_u}\ell(g(x_i^u),+1)- \frac{\pi_n}{n_n} \sum_{i=1}^{n_n}\ell(g(x_i^n),+1) \big \} + \lambda\mbfR(w) \Big\} ,
\end{split}
\end{equation}
where $\mbfR$ is some regularizer, and $\lambda\geq 0 $ is regularization parameter.  And 
 when $\Rts$ in \eqref{risk_sem_1} is used, the corresponding optimization problem is
\begin{equation}
\label{opt_2_g}
\begin{split}
&\min_{w} \Big\{  \frac{a}{n_u} \sum_{i=1}^{n_u}\ell(g(x_i^u),+1) +  \frac{(1-a) \pi_p}{n_p} \sum_{i=1}^{n_p}\ell(g(x_i^p),+1) \\
&\qquad\qquad+  \frac{\pi_n}{n_n} \sum_{i=1}^{n_n}\ell(g(x_i^n),-1) -  \frac{a\pi_n}{n_n} \sum_{i=1}^{n_n}\ell(g(x_i^n),+1) + \lambda\mbfR(w) \Big\}.
\end{split}
\end{equation}

Unfortunately, the objective of  \eqref{opt_2_g} is not guaranteed to be nonnegative due to the negative term $- \frac{a\pi_n}{n_n} \sum_{i=1}^{n_n}\ell(g(x_i^n),+1)$. As pointed out by \cite{Kiryo_NIPS2017}, this can lead to serious overfitting problems. The following theorem provides methods to choose the regularization parameters such that the nonnegativity of the objective of \eqref{opt_2_g} is guaranteed.
\begin{theorem}
    \label{thrm:nonnegative}
 Suppose there exist positive constants $b_1$, $b_2$ and $b_3$ such that 
  \begin{equation} 
  \label{l_condition}
     \ell(t,-1)-\ell(t,+1)\geq -b_1|t|, \quad\text{and}\quad  
   \ell(t,-1) \geq b_2(b_3-|t|).
  \end{equation}
 (In Table \ref{tab:loss} we give examples of loss functions that satisfy \eqref{l_condition},  see their proofs in the supp. material.)
 
(i)  We have 
 \begin{equation*}
\frac{\pi_n}{n_n} \sum_{i=1}^{n_n}\ell(g(x_i^n),-1) -  \frac{a\pi_n}{n_n} \sum_{i=1}^{n_n}\ell(g(x_i^n),+1)\geq (1-a)\pi_n b_2 b_3 -((1-a)b_2 + a  b_1)\frac{\pi_n}{n_n}  \sum_{i=1}^{n_n} |g(x_i^n)|.
\end{equation*}
(ii)  If we choose $\lambda$ and $\mbfR$ such that 
  \begin{equation}
  \label{R_condition}
      \lambda \mbfR(w)\geq ((1-a)b_2 + a  b_1)\frac{\pi_n}{n_n} \sum_{i=1}^{n_n} |g(x_i^n)| -(1-a)\pi_n b_2 b_3
  \end{equation}
   then the objective of \eqref{opt_2_g} is always nonnegative. 
   
(iii) Consider the specific case $g(x)=\langle w, \phi(x)\rangle$, where $\phi:\mbbR^d \to \mbbR^q$ is a feature map transformation. 
The following choices of $\lambda$ and $\mbfR$ satisfy \eqref{R_condition}.
\begin{itemize}[leftmargin=0.5cm]
    \item $\mbfR(w)=\|w\|_2^2$ and $\lambda\geq\frac{( (1-a)b_2 + a b_1)^2\pi_nc^2}{4(1-a) b_2 b_3} $, where $c=\max\{\|\phi(x_i^n)\|_2:i=1,\ldots,n_n\}$ (note that, in practice, we can scale the data to have $c=1$).   
    \item $\mbfR(w)=\|w\|_1$ and $  \lambda\geq  c_{\infty} ((1-a)b_2 b_3 + a  b_1) \pi_n $, where $c_\infty=\max\{\|\phi(x_i^n)\|_\infty:i=1,\ldots,n_n\}$ (in practice, we can scale the data to have $c_\infty=1$).
\end{itemize}
\end{theorem}
 \begin{table}
\caption{Examples of  loss functions satisfying \eqref{l_condition}}
\label{tab:loss}
\begin{center}
\begin{tabular}{cccc}
\toprule 
Name & $\ell(t,y)=\ell(z)$ with $z=ty$ & Bounded & $(b_1,b_2,b_3)$  \\
\midrule
Hinge loss & $\max\{0,1-z\}$ & $\times$ & $(2,1,1)$\\
Double hinge loss& $\max\{0, (1 -z)/2, -z\}$& $\times$ & $(1,1/2,1)$\\
Squared loss &$\frac12 (z-1)^2$& $\times$& $(2,1/2,1/2)$\\ 
Modified Huber loss & $\begin{cases} \max\{0,1-z\}^2 &\mbox{if}\, z\geq -1 \\ - 4z &\mbox{otherwise} \end{cases}  $ &  $\times$ & $(4,1,1/2)$\\
Logistic loss & $\ln (1+ \exp(-z)) $& $\times$ & $(1,1, \ln 2)$\\
Sigmoid loss & $ 1/(1+\exp(z))$ & $\checkmark$ & $(1,1/2,1)$\\
Ramp loss & $ \max\{0, \min\{1,(1- z)/2\}\}$ & $\checkmark$ &  $(1,1/2,1)$\\
\bottomrule
\end{tabular}
\end{center}
\end{table}

We consider both a shallow and deep implementation of the rAD method. In the following, $\pi^e_p$ and $\pi^e_n=1-\pi^e_p$ will denote estimates of the real class-prior probabilities $\pi_p$ and $\pi_n$, respectively.

\hien{\textbf{A shallow rAD method}  \quad 
We plug in $g(x)=\langle w, \phi(x)\rangle$ in \eqref{opt_2_g} (the empirical version of \eqref{risk_sem_1}), where $\phi:\mbbR^d \to \mbbR^q$ is a feature map transformation, and choose the regularization method proposed in Theorem  \ref{thrm:nonnegative} (iii). Specifically, we solve the following minimization problem:
\begin{equation}
\label{opt_2}
\begin{split}
&\min_{w} \Big\{  \frac{a}{n_u} \sum_{i=1}^{n_u}\ell(\wtop \phi(x_i^u),+1) +  \frac{(1-a) \pi^e_p}{n_p} \sum_{i=1}^{n_p}\ell(\wtop \phi(x_i^p),+1) \\
&\qquad\qquad+  \frac{\pi^e_n}{n_n} \sum_{i=1}^{n_n}\ell(\wtop \phi(x_i^n),-1) -  \frac{a\pi^e_n}{n_n} \sum_{i=1}^{n_n}\ell(\wtop \phi(x_i^n),+1) + \lambda\mbfR(w) \Big\}.
\end{split}
\end{equation}
} 

\textbf{A deep rAD method} \quad We plug in  $g(x)=\phi(x;\mcalW)$ in \eqref{opt_1_g} (the empirical version of \eqref{risk_sem_2}), where $\mcalW$ is a set of weights of a deep neural network. Specifically, we train a deep neural network by solving the following optimization problem:
\begin{equation}
\label{opt_1}
\begin{split}
&\min_{\mcalW} \Big\{ \frac{ \pi^e_n}{n_n} \sum_{i=1}^{n_n}\ell(\phi(x_i^n;\mcalW),-1) + \frac{ (1-a) \pi^e_p}{n_p} \sum_{i=1}^{n_p}\ell(\phi(x_i^p;\mcalW),+1) \\
&\qquad\qquad + a \max \big\{ 0, \frac{1}{n_u} \sum_{i=1}^{n_u}\ell(\phi(x_i^u;\mcalW),+1)- \frac{\pi^e_n}{n_n} \sum_{i=1}^{n_n}\ell(\phi(x_i^n;\mcalW),+1) \big \} +\lambda \mbfR(\mcalW) \Big\},
\end{split}
\end{equation}
where $\mbfR$ can be any regularizer. 
Note that we focus on these specific implementations but it is also possible to consider a deep model with \eqref{opt_2_g} or a shallow model with \eqref{opt_1_g}.

\section{Risk bounds}
\label{sec:risk_bound} 
 In this section, we establish the estimation error bound and the excess risk bound for $\hat{g}^1$ and $\hat{g}^2$ which are the empirical risk minimizers obtained by $ \min_{g\in\mathcal G} \Ros(g)$ and $ \min_{g\in\mathcal G} \Rts(g)$, where $ \mcalG$ is a function class. 
 
 Let $g^*$ be the true risk minimizer, that is, $g^*=\arg\min_{g\in\mathcal G} \mcalR(g)$. Throughout this section, we assume that (i)  $\mcalG=\big\{g \,\big | \, \|g\|_{\infty} \leq C_g\big\}$ for some constant $C_g$, and (ii) there exists $C_\ell>0$ such that 
$\sup_{|t|\leq C_g} \max_{y} \ell(t,y) \leq C_\ell$.
It is worth noting that the set of linear classifiers with bounded norms and feature maps is a special case of Condition (i) 
\begin{equation}
\label{G_linear_classifier}
\mcalG=\{ g(x)=\langle w, \phi(x)\rangle_{\mathcal H} \, \big | \, \|w\|_{\mathcal H} \leq C_w, \|\phi(x)\|_{\mathcal H}\leq C_\phi \},
\end{equation}
where $\mathcal H$ is a Hilbert space, $\phi$ is a feature map, and $C_w$ and $C_\phi$ are positive constants. 

Given $g$, $\Rts(g)$ is an unbiased estimator of $\mcalR(g)$ but $\Ros$ is a biased estimator. The following proposition estimates the bias of  $\Ros$ (see Inequality \eqref{consistent_expect}) and shows that, for a fixed $g$, $\Ros (g)$ and $\Rts (g)$ converge to $ \mcalR(g)$ with the rate $ O\big( \frac{\pi_n}{\sqrt{n_n}} + \frac{\pi_p}{\sqrt{n_p}} + \frac{a}{\sqrt{n_u}}\big)$ (see 
Inequality \eqref{consistent_bound_2} and \eqref{consistent_bound}).
 
\begin{proposition}
\label{prop:bias_bound}
Consider a classifier $g$.   Suppose there exists $\rho_g>0$ such that $\mcalR_p^+(g) \geq \rho_g>0$ and denote $\epsilon_g=a\pi_n C_\ell \exp \left( - \frac{2\pi_p^2 \rho_g^2 }{C_\ell^2( 1/n_u + \pi_n^2 /n_n)}\right)$. Then the bias of $\Ros(g)$  satisfies 
\begin{equation}
\label{consistent_expect}
\begin{split}
  0\leq  \expect[\Ros (g)] - \mcalR(g)  \leq \epsilon_g.
\end{split}
\end{equation}
Moreover, for any $\delta>0$,  we have the following inequalities hold with probability at least $1-\delta$
 \begin{equation}
\label{consistent_bound_2}
\begin{split}
  |\Rts (g) - \mcalR(g)| \leq C_\ell \sqrt{\ln(2/\delta)/2} \big(\frac{ (1+a)\pi_n}{\sqrt{n_n}} + \frac{(1-a)\pi_p}{\sqrt{n_p}} + \frac{a}{\sqrt{n_u}} \big),
   \end{split}
\end{equation}
 and 
\begin{equation}
\label{consistent_bound}
\begin{split}
    |\Ros (g) - \mcalR(g)| \leq C_\ell \sqrt{\ln(2/\delta)/2} \big(\frac{ (1+a)\pi_n}{\sqrt{n_n}} + \frac{(1-a)\pi_p}{\sqrt{n_p}} + \frac{a}{\sqrt{n_u}} \big)+ \epsilon_g.
    \end{split}
\end{equation}
\end{proposition}

\textbf{Estimation error bound}\quad
The  Rademacher complexity  of $\mcalG$ for a sample of size $n$ drawn from some distribution $q$ (see e.g., \cite{Mohri2018}) is defined by $\mathfrak R_{n,q}(\mcalG):= \expect_{Z\sim q^n} [\expect_\sigma [\sup_{g\in\mcalG} (\frac{1}{n}\sum_{i=1}^n \sigma_i g(Z_i)) ]],$ where $Z_1,\ldots,Z_n$  are i.i.d random variables following distribution $q$, $Z=(Z_1,\ldots,Z_n)$,   $\sigma_1,\ldots,\sigma_n$  are independent random variables uniformly chosen from $\{-1,1\}$, and $\sigma=(\sigma_1,\ldots,\sigma_n)$.
Similarly to \cite[Theorem 4]{Kiryo_NIPS2017}, we can establish the following estimation error bound for $\hat g^1$.
\begin{theorem}[Estimation error bound for $\hat g^1$]
\label{thrm:error_bound}
We assume that (i) there exists $\rho>0$ such that $\mcalR^+_p(g)\geq \rho $ for all $g\in\mcalG$, (ii) if $g\in\mcalG$ then $-g\in\mcalG$, and (iii) $t\mapsto \ell(t,1)$ and $t\mapsto \ell(t,-1)$ are $L_\ell$-Lipschitz continuous over $\{t: |t|\leq C_g\}$. Denote $\epsilon=a\pi_n C_\ell \exp \left( - \frac{2\pi_p^2 \rho^2 }{C_\ell^2( 1/n_u + \pi_n^2 /n_n)}\right)$. For any $\delta>0$, the following inequality hold  with probability at least $1-\delta$
\begin{equation} 
\label{risk_bound_1}
\begin{split} 
\mcalR(\hat g^1)-\mcalR(g^*)& \leq 8 (1+a) \pi_n L_\ell \mfrakR_{n_n,p_n}(\mcalG) + 8(1-a) \pi_p  L_\ell \mfrakR_{n_p,p_p}(\mcalG) + 8 a  L_\ell \mfrakR_{n_u,p} (\mcalG)+\\
&+ 2 C_\ell \sqrt{\ln(2/\delta)/2} \Big(\frac{ (1+a)\pi_n}{\sqrt{n_n}} + \frac{(1-a) \pi_p}{\sqrt{n_p}}+ \frac{a}{\sqrt{n_u}} \Big)  + 2\epsilon.
\end{split}
\end{equation}
\end{theorem}
By using basic uniform deviation bound \cite{Mohri2018}, the McDiarmid's inequality \cite{McDiarmid1989SurveysIC}, and Talagrand's contraction lemma  \cite{Ledoux1991}, we can prove the following estimation error bound for $\hat g^2$.  
\begin{theorem}[Estimation error bound for $\hat g^2$]
\label{thrm:error_bound_2}
 Assume that   $t\mapsto \ell(t,1)$ and $t\mapsto \ell(t,-1)$ are $L_\ell$-Lipschitz continuous over $\{t: |t|\leq C_g\}$.  For any small $\delta>0$, the following inequality hold  with probability at least $1-\delta$
\begin{equation} 
\label{risk_bound_2}
\begin{split} 
\mcalR(\hat g^2)-\mcalR(g^*)& \leq 4(1-a) \pi_p  L_\ell \mfrakR_{n_p,p_p}(\mcalG) + 4(a+1) \pi_n  L_\ell \mfrakR_{n_n,p_n}(\mcalG) + 4 a  L_\ell \mfrakR_{n_u,p} (\mcalG)+\\
&+ 2 C_\ell \sqrt{\ln(6/\delta)/2} \Big(\frac{(1-a) \pi_p}{\sqrt{n_p}}+\frac{ (1+a)\pi_n}{\sqrt{n_n}} + \frac{a}{\sqrt{n_u}} \Big).
\end{split}
\end{equation}
\end{theorem}
Note that Theorem \ref{thrm:error_bound_2} explicitly states the error bound for $\hat g^2$ with any loss function that satisfies the Lipschitz continuity assumption. The (scaled) ramp loss and the truncated (scaled) squared loss 
 considered in \cite{Sakai_ICML2017} have $L_\ell =1/2$. 

\textbf{Excess risk bound}\quad
The excess risk focuses on the error due to the use of surrogates for the $\zeroone$ loss function. Denote $I^*=\inf_{g\in\mcalF} I(g)$ and $\mcalR^*=\inf_{g\in\mcalF} \mcalR(g)$, where $\mcalF$ is the set of all measurable functions. By using \cite[Theorem 1]{Bartlett2006} (see \eqref{excess_risk} in the supp. material), Theorem \ref{thrm:error_bound}, and Theorem \ref{thrm:error_bound_2}, we can derive the following excess risk bound for $\hat{g}^1$ and $\hat{g}^2$.
\begin{corollary}
If $\ell$ is a classification-calibrated loss (see Definition \ref{def:class_calibrated} in the supp. material), then 
there exists a convex, invertible, and nondecreasing transformation $ \psi_\ell$ with $\psi_\ell(0) = 0$ and  the following inequalities hold with probability at least $1-\delta$
\[ 
\begin{split}
    I(\hat{g}^1) - I^* \leq \psi_{\ell}^{-1} (B_1+ \mcalR (g^*)-\mcalR^*),\quad
      I(\hat{g}^2) - I^* \leq \psi_{\ell}^{-1} (B_2+ \mcalR (g^*)-\mcalR^* ), 
\end{split}
\]
where $B_1$ and $B_2$ are the right hand side of \eqref{risk_bound_1} and \eqref{risk_bound_2}, respectively. 
\end{corollary}


\section{Related work}
\label{sec:related} 
\textbf{AD methods} \quad Outlier detection, novelty detection, and AD are closely related topics. In fact,  these terms have been used interchangeably, and solutions to outlier detection and novelty detection are often used for AD and vice versa. 
 AD methods can be generally classified into three types (i) density-based methods, which estimate the probability distribution of normal instances  \cite{Lecun2006,Li2019MADGANMA,Parzen1962OnJSTOR,Pincus1995}, (ii) reconstruction-based methods, which learn a model that fails to reconstruct anomalous instances but succeeds to reconstruct normal instances  \cite{Dhillon2004,Douglas1974,Hawkins2002,Huang2006,Yan2021LearningDetection}, and (iii) one-class classification methods. We refer the readers to \cite{Ruff2021_review} for a comprehensive review of the three types of AD methods.

\textbf{PU learning methods} \quad Regarding PU learning methods, they
can be classified into three categories:  biased learning, two-step techniques, and  class-prior incorporation. Similarly to one-class classification AD methods, biased PU learning methods make an impractical assumption: they assume/label all unlabeled instances as negative, see e.g.,  \cite{LeeICML2003,Liu_ICDM2003}. Although the PU learning methods using two-step techniques do not have such assumption, they are heuristics since they first identify ``reliable" negative examples and then apply (semi-)supervised learning techniques to the positive labeled instances and the reliable negative instances, see e.g., \cite{LiLiu_IJCAI2003,Chaudhari_ICONLP2012}. To have some theoretical guarantee, the class-prior incorporation methods need to assume that the class priors are known, see e.g.,   \cite{Plessis_NIPS2014,Elkan_KDD2008,Hsieh19c}. We refer the readers to \cite{Bekker2020} and the references therein for more details on the three types of PU learning methods. Methods that rely on risk estimators \cite{Plessis_NIPS2014,Plessis_ICML2015,Kiryo_NIPS2017,Sakai_ICML2017} belong to the third category.


\section{Experiments}
\label{sec:experiments} 
\textbf{A. Experiments with shallow rAD} 

\textit{\textbf{Baseline methods and implementation}} \quad We compare rAD with OC-SVM \cite{Scholkopf2001},  semi-supervised OC-SVM \cite{Mari2010}, and the PU methods using the risk estimator $\hat{\mcalR}_{pu}(g)$ given in \eqref{risk_1}. Note that  $\hat{\mcalR}_{pu}(g)=\hat{\mcalR}^1_{pu}(g)$ given in \eqref{eq:nonconvex_Rpu} if  $\ell$ satisfies the symmetric condition, and $\hat{\mcalR}_{pu}(g)=\hat{\mcalR}^2_{pu}(g)$ given in \eqref{eq:convex_Rpu}  if $\ell$ satisfies the linear-odd condition. We implement rAD and PU methods with 3 losses: squared loss, hinge loss, and modified Huber loss. For rAD, we use $l_2$ regularization and take $\phi(x)=x$ in \eqref{opt_2}, i.e. no kernel is used. We set $a=0.1$ and $\pi^e_p=0.8$ ($\pi^e_n=0.2$) as default values for both the shallow rAD and the PU methods. Note that the real $\pi_n$ of the datasets can be different.  

\textit{\textbf{Datasets}} \quad We test the algorithms on 26 classical  anomaly detection benchmark datasets  from \cite{han2022adbench}, whose $\pi_n$ ranges from 0.02 to 0.4. The real $\pi_n$ of the datasets  are given in the first column of Table \ref{tab:shallow} (and Table \ref{tab:shallow_2} in the supp. material). We randomly split each dataset 30 times into train and test data with a ratio of 7:3, i.e. we have 30 trials for each dataset. Then, for each trial, we randomly select $5\%$ of the train data to make the labeled data and keep the remaining $95\%$ as unlabeled data.

\textit{\textbf{Experimental results}}\quad In Table \ref{tab:shallow}, we report the mean and standard error (SE) of the AUC (area under the ROC curve) over 30 trials of 9 benchmark datasets. The results of the 17 remaining datasets are given in Table \ref{tab:shallow_2} in the supp. material. We observe that, on average, rAD outperforms the PU methods and OC-SVM methods. The difference between the AUC of rAD and that of PU is large on the datasets with $\pi_n\leq 0.2$ but it is small when $\pi_n$ is larger. We also notice that rAD with modified Huber loss often gives better results than rAD with square loss and hinge loss.
 
\begin{table}\captionsetup{font=small}
\caption{Mean (and SE $\times 10^2$) of the AUC over 30 trials. The best means are highlighted in bold. $d$, $n$, and $\pi_n$ denote the feature dimension, the sample size of the dataset, and the ratio of negative samples in the dataset.} 
    \label{tab:shallow}
\centering
    \adjustbox{scale=0.7}{%
        \begin{tabular}{ccccccccc}
    \toprule
   \multirow{2}{*}{\thead{dataset \\$(d,n,\pi_n)$}}  & \multicolumn{3}{c}{rAD} &  \multicolumn{3}{c}{PU} & \multirow{2}{*}{OC-SVM} & \multirow{2}{*}{\thead{semi-\\OC-SVM}}\\
   \cmidrule{2-7}
        & square & hinge & m-Huber & square & hinge & m-Huber & &  \\
              \midrule
                 \thead{pendigits\\  (16, 6870, 0.02)} &   \textbf{0.98}(0.13) &\textbf{0.98}(0.13) &\textbf{0.98}(0.13) & 0.78(3.90) &0.78(3.94) &0.78(3.84) & 0.87(0.25) & 0.77(2.74) 
\\ \addlinespace[-0.7ex]
     \midrule   \thead{mammography\\  (6, 11 183, 0.02)} &   \textbf{0.91}(0.35) &\textbf{0.91}(0.35) &\textbf{0.91}(0.34) & 0.84(2.53) &0.84(2.54) &0.84(2.54) & 0.76(0.48) & 0.58(2.86)  
\\ \addlinespace[-0.7ex]
     \midrule
    \thead{optdigits\\(64, 5216, 0.03)}  & 0.996(0.06) &0.996(0.07) &\textbf{0.997}(0.05) & 0.80(1.97) &0.79(2.02) &0.80(1.98) & 0.48(0.64) & 0.78(2.44)  
\\\addlinespace[-0.7ex]   
     \midrule
     \thead{Stamps\\(9, 340, 0.09)} & \textbf{0.82}(3.56) &0.81(4.08) &0.80(4.14) & 0.70(4.17) &0.74(4.26) &0.66(4.47) & 0.63(1.67) & 0.74(3.37)  
\\\addlinespace[-0.7ex]
     \midrule  
      \thead{cardio\\(21, 1831, 0.10)} &   0.92(1.65) &0.88(1.73) &\textbf{0.93}(1.58) & 0.82(2.34) &0.80(2.46) &0.83(2.23) & 0.86(0.44) & 0.79(1.97)  
   \\\addlinespace[-0.7ex]
    \midrule
    \thead{InternetAds\\(1555, 1966, 0.19)}  & 0.76(2.90) &0.86(0.38) &\textbf{0.87}(0.50) & 0.58(3.54) &0.78(0.51) &0.78(0.56) & 0.60(0.50) & 0.65(1.14)  
\\\addlinespace[-0.7ex]
   \midrule    
    \thead{Cardiotocography\\(21, 2114, 0.22)}  & 0.89(0.92) &0.87(1.10) &\textbf{0.90}(0.87) & 0.82(1.40) &0.80(1.64) &0.83(1.25) & 0.74(0.35) & 0.82(0.79)  
\\\addlinespace[-0.7ex]
   \midrule
        \thead{magic.gamma\\(10, 19 020, 0.35)} & \textbf{0.78}(0.51) &0.77(0.53) &\textbf{0.78}(0.50) & 0.77(0.66) &0.77(0.68) &0.77(0.65) & 0.56(0.18) & 0.54(0.39)  
\\\addlinespace[-0.7ex]
    \midrule
    \thead{SpamBase\\(57, 4207, 0.40)}  & \textbf{0.94}(0.14) &\textbf{0.94}(0.14) &\textbf{0.94}(0.13) & 0.93(0.25) &0.93(0.23) &0.92(0.27) & 0.54(0.40) & 0.63(0.84)  
\\
    \bottomrule
    \end{tabular}%
    }
    \end{table}
    
 \textit{\textbf{Sensitivity analysis for $\pi_p^e$}} \quad  With $a=0.1$, we run shallow rAD on the 30 trials for $\pi_p^e  \in \{1-\pi_n, 0.9, 0.7,0.6\}$ ( when $\pi_p^e=1-\pi_n$, no approximation is made). The results are reported in Table \ref{tab:shallow_3} in the supp. material. From Table \ref{tab:shallow}-- \ref{tab:shallow_3}, we can see that we can obtain good results even if $\pi_p^e$ is different from $\pi_p$. In fact, with $a=0.1$, we get worse AUC means when $\pi_p^e$ is close to $\pi_p$. The combination $(a,\pi_p^e)=(0.1, 0.8)$ or $(a,\pi_p^e)=(0.1, 0.7)$ seem to be good choices across the datasets. Compared to the other two losses, we found the modified Huber loss to be robust to the values of $\pi_p^e$.

\textit{\textbf{Sensitivity analysis for $a$}} \quad   We run shallow rAD (with fixed $\pi_p^e =0.8$) on the 30 trials of each dataset for $a  \in \{0.3, 0.7,0.9\}$. The results are reported in Table \ref{tab:shallow_4} in the supp. material. From  Table \ref{tab:shallow}, \ref{tab:shallow_2} and \ref{tab:shallow_4}, we can see that the AUC means do not decrease significantly when we increase $a$ (except for the dataset InternetAds). Hence, shallow rAD with $\pi_p^e=0.8$ is also robust to different values of $a$.

\textbf{B. Experiments with deep rAD} \quad 


\textit{\textbf{Baseline methods and implementation}}\quad We compare deep rAD with the deep semi-supervised AD method (deep SAD) \cite{Ruff2020Deep}) and the  PU   learning method with nonnegative risk estimator and sigmoid loss (nnPU) \cite{Kiryo_NIPS2017}). For deep SAD and nnPU, we use default hyperparameter settings and network architectures  as in their original implementation by the authors. For deep rAD, we use the same network architectures as deep SAD and we use ADAM to solve the optimization problem in \eqref{opt_1}.   We implement 4 losses for deep rAD: squared loss, sigmoid loss, logistic loss, and modified Huber loss.
We set $a=0.1$ and $\pi^e_p=0.8$ (thus $\pi_n^e=0.2$) as default values for deep rAD.

\textit{\textbf{Datasets}}\quad
 We test the algorithms on 3 benchmark datasets: MNIST, Fashion-MNIST, and CIFAR-10 (all have 10 classes).  We use AD setups following previous works \cite{Chalapathy2018,Ruff2020Deep}: for each $\pi_n \in \{0.01, 0.05, 0.1,  0.2\}$, we set one of the ten classes to be a positive class, letting the remaining nine classes be anomalies and maintaining the ratio between normal instances and anomaly instances such that the setup has the required $\pi_n$ (so we have 10 setups corresponding to 10 classes). We note that the anomalous data in our generation process can originate from more than one of the nine classes (unlike in the setup of deep SAD where the anomaly is only from one of the nine classes).  For each  $\pi_n$, we repeat this generation process 2 times to get 20 AD setups (or 20 trials). Then, in each trial, we randomly choose $\gamma_l$ (with  $\gamma_l\in \{0.05, 0.1, 0.2 \}$) portion of the train data to be labeled and keep the remaining $(1-\gamma_l)$ portion as unlabeled data. Note that we make the labeled data for nnPU only from normal instances. To make labeled data for deep rAD and deep SAD,  $(1-\pi_n)$ portion is taken from the nnPU labeled data (which contain only normal instances), and the remaining $\pi_n$ portion is taken from the anomalous instances. Hence, the number of labeled anomalous instances for deep rAD and deep SAD is about $(\gamma_l\times\pi_n)$ portion of the train data. 
 
\begin{figure}[ht]
  \centering
  \begin{minipage}[b]{0.54\textwidth}
    \centering    \includegraphics[width=\textwidth]{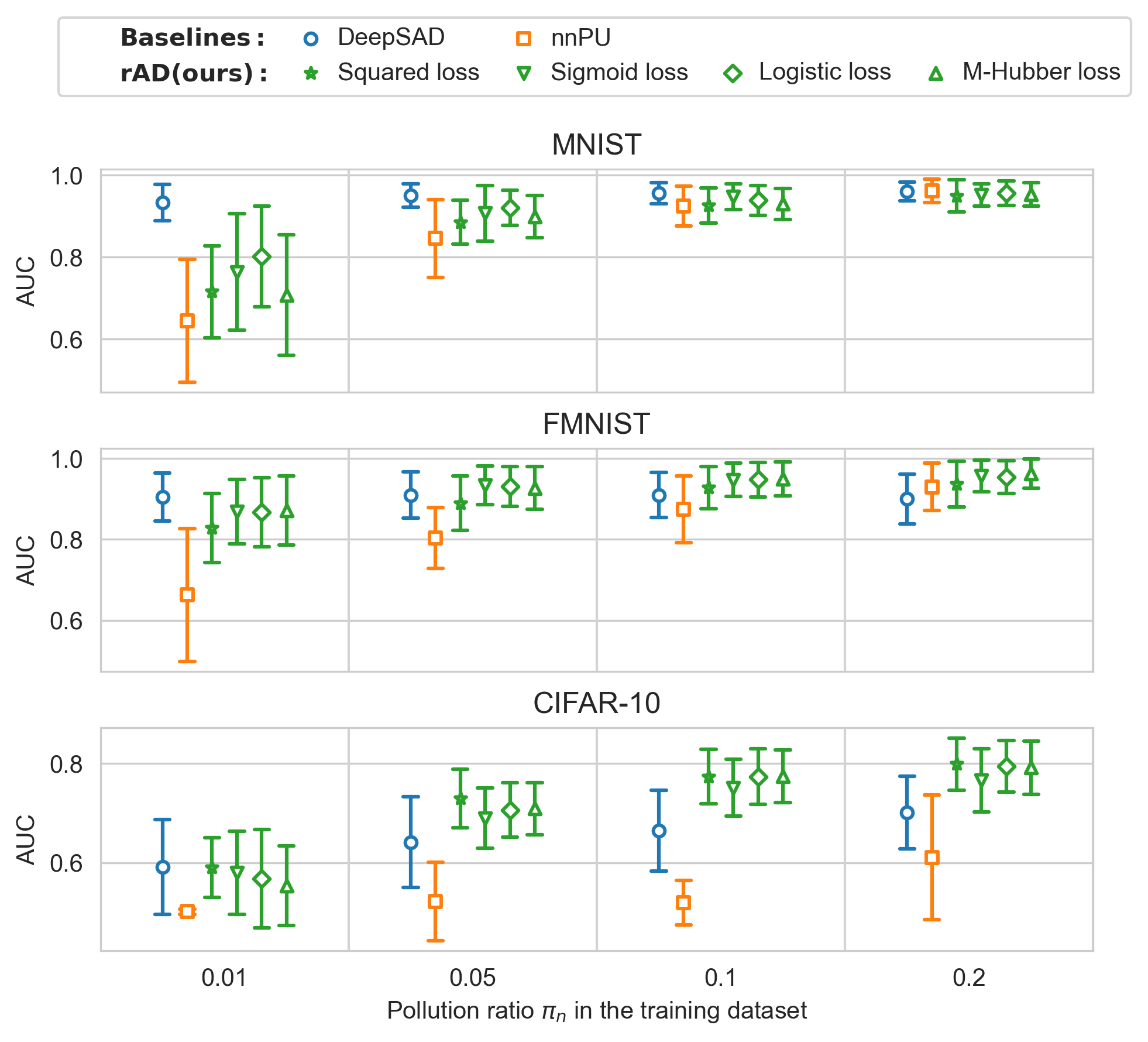}
    \caption{AUC mean and std over 20 trials with various $\pi_n$ and default $\gamma_l = 0.05$}
    \label{fig:results_gamma0.05}
  \end{minipage}
  \hfill
  \begin{minipage}[b]{0.45\textwidth}
    \centering    \includegraphics[width=\textwidth]{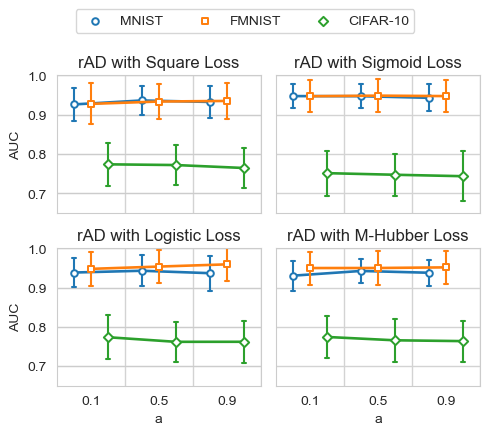}
    \caption{AUC mean and std over 20 trials at various $a$ for the datasets with  $\gamma_l = 0.05$ and $\pi_n = 0.1$}
    \label{fig:results_a}
  \end{minipage}
 \end{figure}

\textit{\textbf{Experiment results}}\quad In Figure \ref{fig:results_gamma0.05}, we report the mean and standard deviation (std) of the AUC over 20 trials on the datasets with increasing pollution ratio $\pi_n$ and default $\gamma_l = 0.05$. The results for $\gamma_l \in \{0.1, 0.2\}$ are given in Figures  \ref{fig:results_gamma0.1} and  \ref{fig:results_gamma0.2} in the supp. material. Figures \ref{fig:results_gamma0.05}, \ref{fig:results_gamma0.1} and \ref{fig:results_gamma0.2} show that, on average, deep rAD methods provide better AUC than deep SAD and nnPU on CIFAR-10. The AUC difference is significant when $\pi_n$ is increased  (deep rAD and deep SAD have similar performance when $\pi_n=0.01$). On FMNIST, deep rAD methods are still better than the others when $\pi_n$ is increased but the AUC improvement is small. On MNIST, deep SAD is best, and deep rAD catch up with it when either $\pi_n$ or $\gamma_l$ is increased. Deep rAD with quadratic loss underperforms the other rAD methods on MNIST and FMNIST. On average, deep rAD with logistic loss performs best among the rAD methods. It is also interesting to note that in the presence of anomalies from multiple classes, the performance of deep SAD degrades over the performance reported in \cite{Ruff2020Deep}. The degradation is more severe for CIFAR-10.

To observe the impact of the amount of labeled data, we report the results for the datasets with $\pi_n=0.1$ and $\gamma_l\in \{0.05, 0.1, 0.2\}$ in Figure \ref{fig:results_difgamma} in the supp. material. We observe that all the methods improve when we increase $\gamma_l$.  From $\gamma_l=0.05$ to $\gamma_l=0.1$ (i.e., 5\% more labeled data), deep rAD methods show a significant improvement in performance.

\textit{\textbf{Sensitivity analysis for  $\pi^e_p$}}\quad 
We run deep rAD with  $a=0.1$ on the 20 trials of each dataset for $\pi_p^e\in \{1-\pi_n, 0.9, 0.7, \pi_n\}$ (when $\pi_p^e=1-\pi_n$, it is an exact estimation of $\pi_p$, and when $\pi_p^e=\pi_n$, we can say $\pi_p^e$ is a bad estimation of $\pi_p$). We report the result in Table \ref{tab:deep_1} in the supp. material. 
Again, we see that $\pi_p^e$ is not necessarily a precise estimation of $\pi_p$; and $(a,\pi_p^e)=(0.1, 0.8)$ or $(a,\pi_p^e)=(0.1, 0.7)$ are good settings. These results are consistent with the results of shallow rAD. 

\textit{\textbf{Sensitivity analysis for $a$}}\quad We fix $\pi_p^e=0.8$ and run deep rAD with additional values of $a\in\{0.5, 0.9\}$ ($a=0.1$ is the default setting). We report the results for the datasets with $\pi_n=0.1$ and $\gamma_l = 0.05$ in Figure \ref{fig:results_a}. 
The results for the datasets with other values of $\pi_n$ and $\gamma_l$ are given in the supp. material. We observe that on CIFAR-10, AUC decreases when $a$ is increased; however, the difference is not significant. On FMNIST and MNIST, deep rAD with $\pi_p^e=0.8$ is quite robust to the change of $a$.

\section{Limitations}
\label{sec:limitation}

Although the experiments have shown that our rAD methods are quite robust to the changes of the parameters $a$ and $\pi_p^e$, we still have to tune them to obtain the best AD performance. On the theoretical side, although the risk bounds are established for the proposed risk minimizers in Section \ref{sec:risk_bound}, we still need the assumption that $\pi_p$ and $\pi_n$ are known in advance, which is a limitation.

 \section{Conclusion}
 \label{sec:conclusion}

With semi-supervised classification based on risk estimators, we have introduced  a shallow AD method equipped with suitable regularization as well as a deep AD method. Theoretically, we have established the estimation error bounds and the excess risk bounds for the two risk minimizers. 
Empirically, the shallow AD methods show significant improvement over the baseline methods while the deep AD methods compete favorably with the baselines. Let us conclude the paper by giving some possible future research directions that address the limitation given in Section \ref{sec:limitation}. One possible research direction  is to develop a method that can learn the best combination of $(a,\pi_p^e)$ from the available data. On the other hand, our experiments have shown that using $a=0.1$, precise estimation of $\pi_p$ and $\pi_n$ are not necessarily needed to obtain good accuracy in terms of AUC. Hence, another  possible research direction would be to study the theoretical bounds of the risk minimizers with $\pi_p$ and $\pi_n$ replaced by some estimates. 



\clearpage 
\bibliography{biblio}
\bibliographystyle{tmlr}
\if\supm1
\newpage 
\appendix
\begin{center}\large{\textbf{
Supplementary material}}\end{center}

\section{Technical proofs}

\subsection{Proof of Theorem \ref{thrm:nonnegative}}
(i)
We have 
\[
\begin{split}
   & \frac{\pi_n}{n_n} \sum_{i=1}^{n_n}\ell(g(x_i^n),-1) -  \frac{a\pi_n}{n_n} \sum_{i=1}^{n_n}\ell(g(x_i^n),+1)\\
    &=(1-a)\frac{\pi_n}{n_n} \sum_{i=1}^{n_n}\ell(g(x_i^n),-1)+  \frac{a\pi_n}{n_n}\sum_{i=1}^{n_n}(\ell(g(x_i^n),-1)-\ell(g(x_i^n),+1)) \\
    &\geq (1-a)\frac{\pi_n}{n_n} \sum_{i=1}^{n_n}b_2(b_3-|g(x_i^n)|) -  a\frac{\pi_n}{n_n} \sum_{i=1}^{n_n}  b_1 |g(x_i^n)|  \\
    &=(1-a)\pi_n b_2 b_3-((1-a)b_2 + a  b_1)\frac{\pi_n}{n_n}  \sum_{i=1}^{n_n} |g(x_i^n)|  . 
\end{split}
\]

(ii) We have \eqref{R_condition} is a direct consequence of Theorem \ref{thrm:nonnegative}(i). 

(iii) Considering the first case,   
$\lambda\geq\frac{( (1-a)b_2 + a b_1)^2\pi_nc^2}{4(1-a) b_2 b_3} $ and $\mbfR(w)=\|w\|_2^2$,  we have     
\[
\begin{split}
     \lambda \mbfR(w) +  (1-a)\pi_n b_2 b_3& \stackrel{\rm(a)}{\geq}   \frac{( (1-a)b_2 + a b_1)^2}{4(1-a) b_2 b_3}   \frac{\pi_n}{n_n^2}(\sum_{i=1}^{n_n} |g(x_i^n)|)^2 +  (1-a)\pi_n b_2 b_3 \\
    &\stackrel{\rm(b)}{\geq} ((1-a)b_2 + a  b_1)\frac{\pi_n}{n_n}  \sum_{i=1}^{n_n} |g(x_i^n)|,
\end{split}
\]
where in (a) we used the property that $ |g(x_i^n)|=|\langle w, \phi(x_i^n)\rangle | \leq c \|w\|_2$, and in (b) we used the inequality $ u + v\geq 2\sqrt{uv}$ for all nonnegative $u$ and $v$. 

Consider the second case, $\mbfR(w)=\|w\|_1$ and $ \lambda\geq  c_{\infty} ((1-a)b_2 + a  b_1) 
 \pi_n$. Note that $|g(x_i^n)|=|\langle w, \phi(x_i^n)\rangle | \leq c_{\infty} \|w\|_1$.  Hence, we have 
\[
\begin{split}
     \lambda \mbfR(w) +  (1-a)\pi_n b_2 b_3& >   c_{\infty} ((1-a)b_2 + a  b_1) 
 \pi_n \|w\|_1\geq  ((1-a)b_2 + a  b_1)  \frac{\pi_n}{n_n}  \sum_{i=1}^{n_n} |g(x_i^n)|. 
\end{split}
\]
\paragraph{Derivation of $b_1$, $b_2$ and $b_3$ in Table \ref{tab:loss} } ~ 

\textbf{Hinge loss.} We have 
\[\begin{split}
    \ell(t,-1)-\ell(t,+1)& = \max\{0,1+t\}-\max\{0,1-t\} =\begin{cases}
     t-1   &\mbox{if}\, t<-1, \\ 
      2t  &\mbox{if}\, -1\leq t \leq 1, \\
       1+t &\mbox{if}\, t>1,
    \end{cases}\\& \geq -2|t|,  
\end{split} 
\]
and  
\[
\begin{split}
   \ell(t,-1) -b_2(1-|t|)=   \max\{0,1+t\} -b_2(1-|t|) =\begin{cases}
   - b_2(1+t)   &\mbox{if}\, t<-1, \\ 
      t+1-b_2-b_2t  &\mbox{if}\, -1\leq t \leq 0, \\
       1+t-b_2+b_2t &\mbox{if}\, t>0,
    \end{cases}
\end{split}
\]
Hence, the hinge loss with $b_1=2$, $b_2=1$  and  $b_3=1$ satisfies \eqref{l_condition}.

\textbf{Double hinge loss.} Similarly, we have
\[\begin{split}
    \ell(t,-1)-\ell(t,+1)& = \max\{0,(1+t)/2,t\}-\max\{0,(1-t)/2,-t\} =t
    \geq -|t|,  
\end{split} 
\]
and  
\[
\begin{split}
   \ell(t,-1) -b_2(1-|t|)&=   \max\{0,(1+t)/2,t\} -b_2(1-|t|) \\&=\begin{cases}
    -b_2(1+t)   &\mbox{if}\, t<-1, \\ 
      t/2+1/2-b_2-b_2t  &\mbox{if}\, -1\leq t \leq 0, \\
       t/2+1/2-b_2+b_2t &\mbox{if}\, 0<t\leq 1, 
       \\
       t-b_2+b_2t &\mbox{if}\, t>1.
    \end{cases}
\end{split}
\]
Hence, the double hinge loss with $b_1=1$, $b_2=1/2$ and $b_3=1$ satisfies \eqref{l_condition}.

\textbf{Square loss.} We have 
\[\begin{split}
    \ell(t,-1)-\ell(t,+1) = \frac12(t+1)^2- \frac12(t-1)^2 = 2t \geq -2|t|.
\end{split} 
\]
Note that when $|t|>1$ we have $(1/2-|t|)<0$, which implies $\ell(t,-1) -1/2(1/2-|t|)>0$. Considering $|t|\leq 1$, when $b_2=1/2$ and $b_3=1/2$, we have 
\[
\begin{split}
   \ell(t,-1) - 1/2( 1/2-|t|)&= 1/2(t+1)^2- 1/2(1/2-|t|)= \begin{cases} t^2 +3/2t +1/4 &\mbox{if}\, 0\leq t\leq 1 \\   t^2 +1/2t +1/4&\mbox{if} \, -1\leq t\leq 0. \end{cases}
   \end{split}
\]
Hence,  the square loss with$b_1=2$, $b_2=1/2$, $b_3=1/2$ satisfies \eqref{l_condition}.4

\textbf{Modified Huber loss.} We have 
\[\begin{split}
    \ell(t,-1)-\ell(t,+1) &= \begin{cases} \max\{0,1+t\}^2 &\mbox{if}\, t\leq 1 \\  4t &\mbox{if} \, t>1 \end{cases}  - \begin{cases} \max\{0,1-t\}^2 &\mbox{if}\, t\geq -1 \\ - 4t &\mbox{if}\, t<-1 \end{cases}\\
    &=4t \geq -4 |t|.
\end{split} 
\]
Considering $|t|\leq 1$, when $b_2=1$, $b_3=1/2$, we have
\[
\begin{split}
   \ell(t,-1) -(1/2-|t|)&= \begin{cases} \max\{0,1+t\}^2 &\mbox{if}\, t\leq 1 \\  4t &\mbox{if} \, t>1  \end{cases} -b_2(1/2-|t|) \\
   &= \begin{cases}  t^2 + 1/2 + t &\mbox{if}\, -1\leq t\leq 0
   \\  t^2 + 5/2 t + 1/2  &\mbox{if} \, 0<t \leq 1  \end{cases}.
   \end{split}
\]
 Hence the modified Huber loss with $b_1=4$, $b_2=1$ and $b_3=1/2$ satisfies \eqref{l_condition}.

 \textbf{Logistic loss.} We have 
\[\begin{split}
    \ell(t,-1)-\ell(t,+1)=t \geq - |t|.
\end{split} 
\]
When $t\geq 0$ then $\ln(1+\exp(t))\geq \ln 2 = b_3 \geq b_3-|t|$. When $ t\leq 0$ we have 
 \[\begin{split}
    \ell(t,-1)-b_2(b_3-|t|)&= \ln(1+\exp(t))-(\ln2+t)\\&=\ln\big(\frac{1+\exp(t)}{2\exp(t)} \big) \geq \ln 1=0. 
\end{split} 
\]
Hence the logistic loss with $b_1=1$, $b_2=1$ and $b_3=\ln2$ satisfies \eqref{l_condition}.

\textbf{Sigmoid loss.} 
When $t> 0$, we have $\ell(t,-1)=\frac{1}{1+\exp(-t)} \geq 1/2 \geq b_2(1-|t|)$. For $t\leq0$, we have 
 \[\begin{split}\ell(t,-1) -b_2(1-|t|) = \frac{1}{1+\exp(-t)} -1/2(1+t)=\frac{1-1/2(1+\exp(-t))(1+t)}{1+\exp(-t)}.
 \end{split} 
\]
Note that the function $t\mapsto 1/2(1+\exp(-t))(1+t)$ is an increasing function on $(-\infty,0]$ and its maximum value on $(-\infty,0]$ is 1. Hence $\ell(t,-1)\geq 1/2(1-|t|)$. 
On the other hand, we have 
\[\begin{split}
    \ell(t,-1)-\ell(t,+1)= 2\ell(t,-1)-1 \geq - |t|.
\end{split} 
\]
Hence, the sigmoid loss with $b_1=1$, $b_2=1/2$ and $b_3=1$ satisfies \eqref{l_condition}.

\textbf{Ramp loss.} We have 
\[\begin{split}
    \ell(t,-1)- b_2(b_3-|t|)&=  \max\{0, \min\{1,(1+t)/2\}\} -  1/2(1-|t|)\\
    &=\begin{cases} -1/2(1+t) &\mbox{if}\,  t\leq -1
   \\  0 &\mbox{if}\, -1\leq t\leq 0
   \\  t &\mbox{if} \, 0<t \leq 1  \\ 
   1/2+1/2t &\mbox{if} \, t\geq 1.
   \end{cases}
\end{split} 
\]
Hence, $\ell(t,-1)\geq 1/2(1-|t|)$.  On the other hand, we have 
\[\begin{split}
    \ell(t,-1)-\ell(t,+1)= 2\ell(t,-1)-1 \geq - |t|.
\end{split} 
\]
Hence, the ramp loss with $b_1=1$, $b_2=1/2$ and $b_3=1$ satisfies \eqref{l_condition}.

\subsection{Proof of Proposition \ref{prop:bias_bound}}
\paragraph{Proof of Inequality \eqref{consistent_expect}}
~

Note that $\expect[\Rts (g)] =\mcalR(g)$. 
Considering $\Ros (g)$, we have 
$ \Ros (g) = \Rts (g)$ on $$ \mcalM^+(g):=\{(\mcalN,\mcalU):  \Rup(g)-\pi_n \Rnp(g) \geq 0\}.$$ 
Denote $ \mcalM^-(g):=\{(\mcalN,\mcalU):  \Rup(g)-\pi_n \Rnp(g) < 0\}$.   We have
\begin{equation}
\label{proof1:temp1}
\begin{split}
\expect[\Ros (g)] - \mcalR(g) &= \expect[\Ros (g) - \Rts (g)]\\
&=\int_{(\mcalN,\mcalU) \in  \mcalM^-(g)} \big(\Ros (g) - \Rts (g) \big) dF(\mcalN,\mcalU)\\
&=\int_{(\mcalN,\mcalU) \in  \mcalM^-(g)} a\big(\pi_n \Rnp(g)- \Rup(g) \big)  dF(\mcalN,\mcalU) \qquad (\ref{proof1:temp1} a)\\
&\leq \sup_{(\mcalN,\mcalU) \in  \mcalM^-(g)} a\big(\pi_n \Rnp(g)- \Rup(g) \big)\int_{(\mcalN,\mcalU) \in  \mcalM^-(g)} dF(\mcalN,\mcalU)\\
&=a \sup_{(\mcalN,\mcalU) \in  \mcalM^-(g)} \big(\pi_n \Rnp(g)- \Rup(g) \big) {\rm Pr} (\mcalM^-(g))
\\&\leq a\pi_n C_\ell {\rm Pr} (\mcalM^-(g)).
\end{split}
\end{equation}
From (\ref{proof1:temp1}a) we have $\expect[\Ros (g)] - \mcalR(g) \geq 0$. On the other hand, 
\[
\begin{split}
{\rm Pr} (\mcalM^-(g))& = {\rm Pr}\big(\Rup(g)-\pi_n \Rnp(g) < 0\big)\\
&\leq {\rm Pr}\big(\Rup(g)-\pi_n \Rnp(g) \leq  \pi_p \mcalR_p^+(g)-\pi_p\rho_g\big)\\
&={\rm Pr}\big(\pi_p \mcalR_p^+(g) - (\Rup(g)-\pi_n \Rnp(g)) \geq \pi_p\rho_g\big)\\
&\leq \exp\left (-\frac{2( \pi_p\rho_g)^2}{n_u( C_\ell/n_u)^2 +  n_n (\pi_n C_\ell/n_n)^2}\right)\\
&=\exp \left( - \frac{2\pi_p^2 \rho_g^2 }{C_\ell^2( 1/n_u + \pi_n^2 /n_n)}\right),
\end{split}
\]
where we have used McDiarmid’s inequality for the last inequality. Therefore, from \eqref{proof1:temp1} we have 
\begin{equation}
    \label{proof1:temp2}
\begin{split}
    \expect[\Ros (g)] - \mcalR(g)  \leq a\pi_n C_\ell \exp \left( - \frac{2\pi_p^2 \rho_g^2 }{C_\ell^2( 1/n_u + \pi_n^2 /n_n)}\right).
\end{split}
\end{equation}
\paragraph{Proof of Inequality \eqref{consistent_bound_2} and   \eqref{consistent_bound}}    
 If an $x_i^n$ is changed then the change of $\Ros (g)$ would be no more than $\pi_n(a+1) C_\ell/n_n $.   If an $x_i^u$ is changed then the change of $\Ros (g)$ would be no more than $a C_\ell/n_u $. And  if an $x_i^p$ is changed then the change of $\Ros (g)$ would be no more than $(1-a)\pi_p C_\ell/n_p$. 
For any $\delta>0$, let  $$\varepsilon=C_\ell\sqrt{\big(\frac{(1+a)^2\pi_n^2}{n_n }+\frac{(1-a)^2 \pi_p^2}{n_p} + \frac{a^2}{n_u} \big)\ln(2/\delta)/2}.$$
 Applying McDiarmid's inequality, we get 
\begin{equation}
\label{temp1}
\begin{split}
&{\rm Pr} ( |\Ros (g)-\expect[\Ros (g)] |\geq \varepsilon)\\
&  \leq  2\exp\left( -\frac{2\varepsilon^2}{n_n(\pi_n (1+a) C_l/n_n )^2 + n_p((1-a)\pi_p C_l/n_p)^2+ n_u (a C_\ell/n_u )^2} \right)\\
&=\delta.
\end{split}
\end{equation}
Hence, 
$$  |\Ros (g)-\expect[\Ros (g)] |\leq \varepsilon \leq C_\ell \sqrt{\ln(2/\delta)/2} \big(\frac{ (1+a)\pi_n}{\sqrt{n_n}} + \frac{(1-a)\pi_p}{\sqrt{n_p}} + \frac{a}{\sqrt{n_u}} \big)$$ with probability at least $1-\delta$. Together with  \eqref{proof1:temp2} and 
\[
\begin{split}
    |\Ros (g) - \mcalR(g)| \leq |\Ros (g)-\expect[\Ros (g)] | + |\expect[\Ros (g)] - \mcalR(g)|,
\end{split}
\]
we obtain Inequality \eqref{consistent_bound} with probability at least $1-\delta$.

Similarly, by applying  McDiarmid's inequality, we obtain Inequality \eqref{consistent_bound_2} with probability at least $1-\delta$.

\subsection{Proof of Theorem \ref{thrm:error_bound}}
Denote $ \Rnu(g)=\pi_n \Rnn(g) +  a \max\{ 0, \Rup(g)-\pi_n \Rnp(g)  \} $. Note that  
$$ \Ros(g)=(1-a)\pi_p \Rpp + \tilde{\mcalR}_{nu}(g). $$ 
 We have
\begin{equation}
    \label{proof2:temp1}
    \begin{split}
       \mcalR(\hat g^ 1)-\mcalR(g^*) &= \mcalR(\hat g^ 1)-\Ros (\hat g^1) + \Ros (\hat g^1) - \Ros (g^*) + \Ros (g^*) - \mcalR(g^*) \\
       &\stackrel{\rm(a)}{\leq} |\Ros (\hat g^1)-\mcalR(\hat g^ 1)| + |\Ros (g^*) - \mcalR(g^*)|
       \\
       &\leq 2 \supgg |\Ros ( g)-\mcalR( g)|\\
       &\leq 2\big( \supgg \big|\Ros ( g)-\expect[\Ros ( g)]\big| + \supgg \big|\expect[\Ros ( g)]- \mcalR( g)\big|\big)\\
       &\stackrel{\rm(b)}{\leq}   2 \supgg \big|\Ros ( g)-\expect[\Ros ( g)]\big| + 2\epsilon\\
       &\leq 2 (1-a)\pi_p \supgg\big| \Rpp - \expect[\Rpp]\big| +  2 \supgg \big| \Rnu(g) - \expect[\Rnu(g)  ]\big| + 2\epsilon,
    \end{split}
\end{equation}
where we used $\Ros (\hat g^1) - \Ros (g^*)\leq 0$ for  (a),  and used \eqref{consistent_expect} for (b). 

To obtain a bound for $\supgg \big| \Rnu(g) - \expect[\Rnu(g)  ]\big|$ we adapt the technique of \cite[Theorem 4]{Kiryo_NIPS2017}. 
Note that for a fix $g$, $\expect[\Rnu( g)]$ is a constant. Hence,  if an $x_i^n$, or $x_i^u$ is changed then the change of $\supgg \big|\Rnu ( g)-\expect[\Rnu ( g)]\big|$ would be the supremum of the change of $\Rnu ( g)$.  By applying McDiarmid's inequality to $\supgg \big|\Rnu ( g)-\expect[\Rnu ( g)]\big|$, we have 
\begin{equation}
    \label{proof2:temp6}
\begin{split}
&\supgg \big|\Rnu ( g)-\expect[\Rnu ( g)]\big| - \expect\big[\supgg \big|\Rnu ( g)-\expect[\Rnu ( g)]\big| \big] \\
 & \quad\leq  C_\ell \sqrt{\ln(2/\delta)/2} \big(\frac{ (1+a)\pi_n}{\sqrt{n_n}} + \frac{a}{\sqrt{n_u}} \big)
\end{split}
\end{equation}
with probability at least $1-\delta/2$. 

Let $(\mcalN',\mcalU')$  be a ghost sample identical to $(\mcalN,\mcalU)$. We have  
\begin{equation}
    \label{sup_bound}
    \begin{split}
  &  \expect\big[\supgg \big|\Rnu ( g)-\expect[\Rnu ( g)]\big| \big] 
  \\& = \expect_{(\mcalN,\mcalU)}\big[\supgg \big|\Rnu ( g;\mcalN,\mcalU)-\expect_{(\mcalN',\mcalU')}[\Rnu ( g;\mcalN',\mcalU')]\big| \big] 
    \\
    &=\expect_{(\mcalN,\mcalU)} \big[\supgg \big|\expect_{(\mcalN',\mcalU')} [\Rnu ( g;\mcalN,\mcalU)- \Rnu ( g;\mcalN',\mcalU')]\big|\big] \\&\leq \expect_{(\mcalN,\mcalU),(\mcalN',\mcalU')} \big[\supgg \big| \Rnu ( g;\mcalN,\mcalU)- \Rnu ( g;\mcalN',\mcalU') \big| \big],
    \end{split}
\end{equation}
where we applied Jensen's inequality. Furthermore, we have
\begin{equation}
\label{proof2:temp2}
\begin{split}
 &   \big| \Rnu ( g;\mcalN,\mcalU)- \Rnu ( g;\mcalN',\mcalU') \big| 
 \\&\leq \pi_n\big|\hat{\mcalR}_n^-(g;\mcalN)-\hat{\mcalR}_n^-(g;\mcalN') \big| \\
 &\qquad+  a\big| \max\{ 0, \Rup(g;\mcalU)-\pi_n \Rnp(g;\mcalN)  \}- \max\{ 0, \Rup(g;\mcalU')-\pi_n \Rnp(g;\mcalN')  \} \big| \\
 &\leq  \pi_n\big|\hat{\mcalR}_n^-(g;\mcalN)-\hat{\mcalR}_n^-(g;\mcalN') \big| \\
 &\qquad+a\big|  \Rup(g;\mcalU) -  \Rup(g;\mcalU')\big| + a\pi_n\big| \Rnp(g;\mcalN)-\Rnp(g;\mcalN') \big|. 
 \end{split}
\end{equation}
Hence, from \eqref{sup_bound} and \eqref{proof2:temp2}, we obtain 
\begin{equation}
    \label{proof2:temp4}
    \begin{split}
      &   \expect\big[\supgg \big|\Rnu ( g)-\expect[\Rnu ( g)]\big| \big]  \\
      &\leq  \pi_n\expect_{\mcalN,\mcalN'}\big[\supgg\big|  \Rnn(g;\mcalN) -  \Rnn(g;\mcalN')\big|  \big] + a \expect_{\mcalU,\mcalU'}\big[\supgg\big|  \Rup(g;\mcalU) -  \Rup(g;\mcalU')\big|  \big]\\
      &\quad  
      + a\pi_n \expect_{\mcalN,\mcalN'}\big[\supgg\big|  \hat\mcalR^+_n(g;\mcalN) -  \hat\mcalR^+_n(g;\mcalN')\big|  \big].
    \end{split}
\end{equation}
Now by using the same technique of \cite[Lemma 5]{Kiryo_NIPS2017}, we can prove that 
\begin{equation}
\label{proof2:temp5}
\begin{split}
    \expect_{\mcalN,\mcalN'}\big[\supgg\big|  \Rnn(g;\mcalN) -  \Rnn(g;\mcalN')\big|  \big]&\leq 4 L_\ell \mfrakR_{n_n,p_n}(\mcalG),\\
    \expect_{\mcalU,\mcalU'}\big[\supgg\big|  \Rup(g;\mcalU) -  \Rup(g;\mcalU')\big|  \big]&\leq 4 L_\ell \mfrakR_{n_u,p} (\mcalG),\\
    \expect_{\mcalN,\mcalN'}\big[\supgg\big|  \hat\mcalR^+_n(g;\mcalN) -  \hat\mcalR^+_n(g;\mcalN')\big|  \big] &\leq 4 L_\ell \mfrakR_{n_n,p_n} (\mcalG).
\end{split}
\end{equation}
For completeness, let us provide the proof in the following. 

Denote $\tilde{\ell}(t,y)=\ell(t,y)-\ell(0,y)$. Then, $\tilde{\ell}(0,y)=0$. Note that $t\mapsto \tilde{\ell}(t,y)$ is also $L_\ell$-Lipschitz continuous over $\{t:|t|\leq C_g\}$. Denote $ \mfrakR'_{n,q}(\mcalG):= \expect_{Z\sim q^n} [\expect_\sigma [\sup_{g\in\mcalG} |\frac{1}{n}\sum_{i=1}^n \sigma_i g(Z_i)| ]]$. We prove the first inequality of \eqref{proof2:temp5}, the others can be proved similarly. We have 
\[
\begin{split}
   & \expect_{\mcalN,\mcalN'}\big[\supgg\big|  \Rnn(g;\mcalN) -  \Rnn(g;\mcalN')\big|  \big]\\
    &=  \expect_{\mcalN,\mcalN'}\big[\supgg\big|\frac{1}{n_n}\sum_{i=1}^{n_n} \ell(g(x_i^n),-1)-\frac{1}{n_n}\sum_{i=1}^{n_n} \ell(g({x'}_i^{n}),-1)\big|  \big]
     \\&=  \expect_{\mcalN,\mcalN'}\big[\supgg\big|\frac{1}{n_n}\sum_{i=1}^{n_n}\big(\tilde{ \ell}(g(x_i^n),-1)- \tilde{\ell}(g({x'}_i^{n}),-1)\big)\big|  \big]
 \\
     &\stackrel{\rm(a)}{=} 
     \expect_{\mcalN,\mcalN',\sigma}\big[ \supgg\big|\frac{1}{n_n}\sum_{i=1}^{n_n}\sigma_i\big(\tilde{ \ell}(g(x_i^n),-1)- \tilde{\ell}(g({x'}_i^{n}),-1)\big)\big|  \big]\\
     &     \leq 2 \mfrakR'_{n_n,p_n}  (\tilde{\ell}(\cdot,-1)\circ\mcalG)\stackrel{\rm(b)}{\leq} 4 L_\ell \mfrakR'_{n_n,p_n}  (\mcalG) \stackrel{\rm(c)}{=} 4L_\ell \mfrakR_{n_n,p_n}  (\mcalG), 
\end{split}
\]
where in (a) we used the property that $\sigma_i$ are independent uniformly distributed random variables taking values in $\{-1,+1\}$, in (b) we use \cite[Theorem 4.12]{Ledoux1991},  and in (c) we use the assumption that both $g$ and $-g$ are in $\mcalG$. 

On the other hand, in a similar manner, we can prove that
\begin{equation}
    \label{proof2:temp3}
\supgg\big| \Rpp - \expect[\Rpp]\big|\leq 4 L_\ell \mfrakR_{n_p,p_p}(\mcalG) + C_\ell\frac{\sqrt{\ln(2/\delta)/2}}{\sqrt{n_p}}
\end{equation}
with probability at least $1-\delta/2$. 
From \eqref{proof2:temp1}, \eqref{proof2:temp6}, \eqref{proof2:temp4}, \eqref{proof2:temp5}, and \eqref{proof2:temp3}, we obtain the result.  

\subsection{Proof of Theorem \ref{thrm:error_bound_2}}

We have
\begin{equation}
    \label{proof3:temp1}
    \begin{split}
        \mcalR(\hat g^2) - \mcalR(g^*)&= \mcalR(\hat g^ 2)-\Rts (\hat g^2) + \Rts (\hat g^2) - \Rts (g^*) + \Rts (g^*) - \mcalR(g^*)\\
        &\stackrel{\rm(a)}{\leq}   \mcalR(\hat g^ 2)-\Rts (\hat g^2)  + \Rts (g^*) - \mcalR(g^*)\\
        &\leq \supgg \big(  \mcalR(g)-\Rts (g)  \big) + \supgg \big( \Rts (g) - \mcalR(g) \big) \\
        & \stackrel{\rm(b)}{=}  \supgg \big(  \expect[\Rts ( g)]-\Rts (g)  \big) + \supgg \big( \Rts (g) -\expect[\Rts ( g)] \big) 
    \end{split}
\end{equation}
where in (a)  we have used $\Rts (\hat g^2) \leq \Rts (g^*) $ and in (b) we have used $\expect[\Rts ( g)]=\mcalR( g)$ given $g$. 

We have 
\begin{equation}
\label{source_ie1}
    \begin{split}
    \supgg \big(  \expect[\Rts ( g)]-\Rts (g)  \big) &\leq a\supgg \big(   \expect[\Rup(g)] -\Rup(g)\big) +  (1-a)\pi_p\supgg \big( \expect[\Rpp(g)]- \Rpp(g) \big) \\&\quad \qquad+ \supgg \big( \expect[\pi_n\Rnn(g) - a\pi_n \Rnp]-(\pi_n\Rnn(g) - a\pi_n \Rnp )  \big),
        \end{split}  
    \end{equation}
    and
 \begin{equation}
 \label{source_ie2}
    \begin{split}
    \supgg \big( \Rts (g) - \expect[\Rts ( g)] \big) &\leq a\supgg \big(  \Rup(g)- \expect[\Rup(g)]\big) +  (1-a)\pi_p\supgg \big(\Rpp(g)- \expect[\Rpp(g)]  \big) \\&\quad \qquad+ \supgg \big( (\pi_n\Rnn(g) - a\pi_n \Rnp )-\expect[\pi_n\Rnn(g) - a\pi_n \Rnp]  \big),
        \end{split}  
    \end{equation}   
Applying McDiarmid's inequality to $\supgg \big(   \expect[\Rup(g)] -\Rup(g)\big)$ we have 
\begin{equation}
    \label{bound_temp_1}
\begin{split}
   \supgg \big(   \expect[\Rup(g)] -\Rup(g)\big) -  \expect \big[ \supgg \big(   \expect[\Rup(g)] -\Rup(g)\big)  ] \leq C_\ell \sqrt{\ln(6/\delta)/2} \frac{1}{\sqrt{n_u}}
\end{split}
\end{equation}
with probability at least $\delta/6$. Moreover, letting $\mcalU'$ be a ghost sample identical to $\mcalU$, we have  
\[
\begin{split}
 \expect \big[ \supgg \big(   \expect[\Rup(g)] -\Rup(g)\big)  ] & =   \expect_{\mcalU} \big[ \supgg \big(   \expect_{\mcalU'}[\Rup(g;\mcalU')] -\Rup(g;\mcalU)\big)  ]\\
 &\stackrel{\rm(a)}{\leq} \expect_{\mcalU,\mcalU'}  \big[ \supgg    \big(  \Rup(g;\mcalU') -\Rup(g;\mcalU)\big)\big]\\
 &=\expect_{\mcalU,\mcalU'}  \big[ \supgg    \big(  \frac{1}{n_u} \sum_{i=1}^{n_u} ( \ell(g({x'}_i^u),+1) -\ell(g(x_i^u),+1))\big)\big]\\
 &\stackrel{\rm(b)}{\leq} \expect_{\mcalU,\mcalU',\sigma}  \big[ \supgg    \big(  \frac{1}{n_u} \sum_{i=1}^{n_u} \sigma_i( \ell(g({x'}_i^u),+1) -\ell(g(x_i^u),+1))\big)\big]\\
 &\leq 2 L_\ell \mfrakR_{n_u,p}(\mcalG).
\end{split}
\]
where we have used the sub-additivity of the supremum in (a) and the property of $\sigma$ in (b). 
Together with \eqref{bound_temp_1} we get 
\begin{equation}
    \label{bound_temp_2}
     \supgg \big(   \expect[\Rup(g)] -\Rup(g)\big) \leq 2 L_\ell \mfrakR_{n_u,p}(\mcalG) + C_\ell \sqrt{\ln(6/\delta)/2} \frac{1}{\sqrt{n_u}}
\end{equation}
with probability at least $\delta/6$. 
 Similarly, we can prove the following 
 inequalities hold with a probability of at least $1-\delta/6$ 
\begin{equation}
     \label{proof3:temp2}
    \begin{split}
    &\supgg \big( \Rup(g)-  \expect[\Rup(g)]\big) \leq 2 L_\ell \mfrakR_{n_u,p}(\mcalG) + C_\ell \sqrt{\ln(6/\delta)/2} \frac{1}{\sqrt{n_u}}, \\
  &   \supgg\big( \Rpp(g) - \expect[\Rpp(g)]\big)\leq 2 L_\ell \mfrakR_{n_p,p_p}(\mcalG) + C_\ell\sqrt{\ln(6/\delta)/2}\frac{1}{\sqrt{n_p}},\\
  &   \supgg\big(  \expect[\Rpp(g)]-\Rpp(g) \big)\leq 2 L_\ell \mfrakR_{n_p,p_p}(\mcalG) + C_\ell\sqrt{\ln(6/\delta)/2}\frac{1}{\sqrt{n_p}},\\
  & \supgg\big (\pi_n\Rnn(g) - a\pi_n \Rnp - \expect[\pi_n\Rnn(g) - a\pi_n \Rnp] \big)\\
     &\qquad\qquad\leq 2 L_\ell (1+a)\pi_n \mfrakR_{n_n,p_n}(\mcalG) + C_\ell (1+a)\pi_n\sqrt{\ln(6/\delta)/2}\frac{1}{\sqrt{n_n}},\\
  & \supgg\big ( \expect[\pi_n\Rnn(g) - a\pi_n \Rnp] -\pi_n\Rnn(g) - a\pi_n \Rnp\big)\\
     &\qquad\qquad\leq 2 L_\ell (1+a)\pi_n \mfrakR_{n_n,p_n}(\mcalG) + C_\ell (1+a)\pi_n\sqrt{\ln(6/\delta)/2}\frac{1}{\sqrt{n_n}}.
    \end{split}
\end{equation}
From \eqref{proof3:temp1}, \eqref{source_ie1}, \eqref{source_ie2}, \eqref{bound_temp_2}, and \eqref{proof3:temp2}, we get the result.

\section{Some definitions}
\begin{definition}
\label{def:class_calibrated}
A loss $\ell$ is said to be classification-calibrated if, for any  $\eta\ne\frac12$, we have $H_\ell^-(\eta) > H_\ell(\eta)$, 
where 
\[
\begin{split} H_\ell(\eta)&=\inf_{\alpha\in \mbbR} (\eta\ell(\alpha,+1) + (1-\eta) \ell(\alpha,-1) ), \\
 H^-_\ell(\eta)&=\inf_{\alpha\in \mbbR: \alpha(\eta-\frac12)\leq 0} (\eta\ell(\alpha,+1) + (1-\eta) \ell(\alpha,-1) )
 \end{split}
\]
\end{definition}
Examples of classification-calibrated loss include the scaled ramp loss,  the hinge loss, and the exponential loss. \cite[Theorem 1]{Bartlett2006} shows that if $\ell$ is a classification-calibrated loss, then there exists a convex,
invertible and nondecreasing transformation $ \psi_\ell$ with $\psi_\ell(0) = 0$ and
$\psi_\ell(I(g) - I^*) \leq  \mcalR(g) - \mcalR^*$,
which implies that 
\begin{equation}
    \label{excess_risk}
    \begin{split}
    I(g) - I^*&\leq \psi_{\ell}^{-1}  (\mcalR(g) - \mcalR^*) =\psi_{\ell}^{-1}  (\mcalR(g) -\mcalR(g^*) + \mcalR (g^*)-\mcalR^*).
    \end{split}
\end{equation}
\section{Additional experiments}

\subsection{Additional experiments for shallow rAD} 
Table \ref{tab:shallow_2} reports the mean and standard error of the AUC of rAD and PU learning methods over 30 trials for the additional benchmark datasets from \cite{han2022adbench}. 

\begin{table}\captionsetup{font=small}
\caption{Mean (and SE $\times 10^2$) of the AUC over 30 trials. The best means are highlighted in bold. $d$, $n$, and $\pi_n$ denote the feature dimension, the sample size of the dataset, and the ratio of negative samples in the dataset.}
     \label{tab:shallow_2}
\begin{center}
    \adjustbox{scale=0.75}{%
        \begin{tabular}{ccccccccc}
    \toprule
   \multirow{2}{*}{\thead{Dataset \\$(d,n,\pi_n)$}}  & \multicolumn{3}{c}{rAD} &  \multicolumn{3}{c}{PU} & \multirow{2}{*}{OC-SVM} & \multirow{2}{*}{\thead{semi-\\OC-SVM}}\\
   \cmidrule{2-7}
        & square & hinge & m-Huber & square & hinge & m-Huber & &  \\
              \midrule
              \thead{satimage-2\\  (36, 5803, 0.01) }&   0.98(0.46) &0.98(0.69) &\textbf{0.99}(0.33) & 0.80(4.10) &0.76(4.03) &0.81(4.12) & \textbf{0.99}(0.09) & 0.55(3.66)  
\\ \addlinespace[-0.7ex]
     \midrule
      \thead{thyroid\\  (6, 3772, 0.02)} &  \textbf{0.995}(0.06) &\textbf{0.995}(0.07) &\textbf{0.995}(0.06) & 0.85(3.57) &0.85(3.64) &0.85(3.5) & 0.92(0.32) & 0.63(4.42)  
\\ \addlinespace[-0.7ex]
     \midrule
         \thead{vowels\\(12, 1456, 0.03)}  & 0.87(1.73) &0.83(1.85) &\textbf{0.88}(1.69) & 0.64(2.29) &0.63(2.22) &0.64(2.36) & 0.72(1.34) & 0.71(3.23)  
\\\addlinespace[-0.7ex]
    \midrule
     \thead{Waveform\\(21, 3443, 0.03)} & 0.84(1.34) &0.82(1.56) &\textbf{0.85}(1.19) & 0.62(3.34) &0.62(3.23) &0.62(3.37) & 0.65(0.61) & 0.72(1.75)  
\\\addlinespace[-0.7ex]
    \midrule
    \thead{CIFAR10-1\\(512, 5263, 0.05)}  & \textbf{0.76}(1.16) &0.75(1.12) &\textbf{0.76}(1.15) & 0.60(1.68) &0.58(1.76) &0.60(1.63) & 0.64(0.41) & 0.70(1.03)
\\\addlinespace[-0.7ex]   
     \midrule
     \thead{SVHN-1\\(512, 10 000, 0.05)} & \textbf{0.84}(0.36) &0.83(0.39) &\textbf{0.84}(0.34) & 0.70(1.31) &0.70(1.40) &0.70(1.27) & 0.66(0.29) & 0.72(0.61)  
  \\\addlinespace[-0.7ex]
     \midrule
    \thead{20news-1\\(768, 2514, 0.05)}  &  0.69(1.67) &0.63(1.21) &\textbf{0.76}(1.20) & 0.57(1.62) &0.54(1.39) &0.59(1.63) & 0.51(0.99) & 0.67(1.49) 
\\\addlinespace[-0.7ex]
     \midrule
      \thead{agnews-1\\(768, 10000, 0.05)} &0.97(0.31) &0.93(0.68) &\textbf{0.98}(0.20) & 0.81(1.04) &0.76(1.24) &0.83(0.91) & 0.75(0.31) & 0.89(0.47)  
\\\addlinespace[-0.7ex]
    \midrule
     \thead{ amazon\\(768, 10000, 0.05)}  & 0.81(0.64) &0.76(0.94) &\textbf{0.83}(0.57) & 0.62(1.37) &0.59(1.39) &0.63(1.31) & 0.55(0.38) & 0.79(0.54)  
\\\addlinespace[-0.7ex]
    \midrule
    \thead{imdb\\(768, 10000, 0.05)}  & 0.82(0.70) &0.78(0.85) &\textbf{0.84}(0.64) & 0.64(1.19) &0.61(1.13) &0.65(1.18) & 0.50(0.26) & 0.77(0.54)  
  \\\addlinespace[-0.7ex]
    \midrule
    \thead{yelp\\(768, 10000, 0.05)} & 0.89(0.63) &0.84(1.07) &\textbf{0.91}(0.47) & 0.68(1.29) &0.64(1.33) &0.70(1.25) & 0.60(0.38) & 0.83(0.54)  
\\\addlinespace[-0.7ex]
     \midrule
     \thead{mnist\\(100, 7603, 0.09)}  & \textbf{0.97}(0.15) &0.96(0.15) &\textbf{0.97}(0.14) & 0.93(5.93) &0.93(5.98) &0.93(5.92) & 0.80(0.26) & 0.86(0.65) 
\\\addlinespace[-0.7ex]
     \midrule
         \thead{campaign
\\(62, 41 188, 0.11)} &  \textbf{0.85}(0.16) &\textbf{0.85}(0.16) &\textbf{0.85}(0.16) & 0.83(0.26) &0.83(0.26) &0.84(0.27) & 0.69(0.13) & 0.77(0.43)  \\\addlinespace[-0.7ex]
\midrule\thead{vertebral\\(6, 240, 0.13)}  & 0.72(2.95) &0.70(2.89) &\textbf{0.73}(3.01) & 0.54(3.15) &0.55(3.46) &0.55(3.16) & 0.47(2.02) & 0.72(2.48)
\\\addlinespace[-0.7ex]
    \midrule
        \thead{landsat\\(36, 6435, 0.21)} &  \textbf{0.74}(0.15) &\textbf{0.74}(0.17) &\textbf{0.74}(0.14) & 0.71(0.79) &0.71(0.80) &0.71(0.79) & 0.36(0.26) & 0.76(0.54) 
\\\addlinespace[-0.7ex]
     \midrule
    \thead{satellite\\(36, 6435, 0.32)} &  \textbf{0.80}(0.22) &\textbf{0.80}(0.22) &\textbf{0.80}(0.22) & \textbf{0.80}(0.34) &0.79(0.32) &\textbf{0.80}(0.36) & 0.54(0.23) & 0.67(0.75)  
\\\addlinespace[-0.7ex]
    \midrule
    \thead{fault\\(27, 1941, 0.35)} &  0.65(0.51) &0.62(0.54) &\textbf{0.66}(0.50) & 0.61(1.01) &0.59(1.02) &0.60(1.00) & 0.53(0.45) & 0.60(0.80)  
\\ 
    \bottomrule
    \end{tabular} %
    }
    \end{center}
\end{table}
Table \ref{tab:shallow_3} reports the mean of the AUC of shallow rAD over the 30 trials for different values of $\pi_p^e$. 

\begin{table}\captionsetup{font=small}
\caption{AUC means of shallow rAD over 30 trials for different $\pi_p^e$. The significant changes in the AUC means are highlighted in bold.}
    \label{tab:shallow_3}
\centering
    \adjustbox{scale=0.75}{%
        \begin{tabular}{ccccccccccccc}
    \toprule
   \multirow{2}{*}{Dataset}  & \multicolumn{4}{c}{square\big/$\pi_p^e$ } &  \multicolumn{4}{c}{hinge\big/$\pi_p^e$ } & \multicolumn{4}{c}{m-Huber\big/$\pi_p^e$} \\
   \cmidrule{2-13}
        & $1-\pi_n$ & 0.9 & 0.7& 0.6 & $1-\pi_n$ & 0.9 & 0.7& 0.6 & $1-\pi_n$ & 0.9 & 0.7& 0.6  \\
              \midrule
              thyroid& 0.98  & 0.995 & 0.996& 0.996& 0.97 & 0.994& 0.996 & 0.996& 0.99 & 0.996&0.996 &0.996
\\
     \midrule
  pendigits &0.96   & 0.98 & 0.98 & 0.98 &0.94&0.98  &0.98 &0.98 &0.97 &0.98 &0.98 &0.98 
\\ 
     \midrule
     Waveform &\textbf{0.74} & 0.82 & 0.84 & 0.84 & \textbf{0.70} &\textbf{0.78} & 0.83 &0.83 &\textbf{0.77} &0.84 &0.85 & 0.85 
\\
    \midrule
   optdigits & 0.96& 0.99 & 0.997& 0.997 &\textbf{ 0.93} & 0.99 &0.997&0.997 & 0.98 &0.996& 0.998 & 0.998
\\  
     \midrule
     Stamps & 0.80 &0.80 &0.82 &0.82 &0.81 &0.81&0.81&0.80 &0.80 &0.80&0.80&0.80
\\
     \midrule
    mnist & 0.96 &0.96 &0.97&0.97 & 0.96 & 0.96 &0.96 &0.96 & 0.97&0.97&0.97&0.97
\\
     \midrule
     cardio & 0.91  &0.91&0.92&0.92&0.87 & 0.88 &0.88 & 0.89 &0.92 & 0.93 & 0.94 & 0.94
\\
    \midrule
    campaign& 0.85 &0.85&0.85&0.85 & 0.85&0.85&0.85&0.85 & 0.85&0.85&0.85&0.85
\\
    \midrule
  InternetAds  &0.77 &0.77&0.70&\textbf{0.60}&  0.86 &0.85&0.86&0.86 & 0.87 &0.87& 0.86 & 0.86
\\
    \midrule
  landsat & 0.74 &0.74&0.74&0.74& 0.74&0.73&0.74&0.74& 0.74&0.74&0.74&0.74
\\
     \midrule
    Cardiotocography & 0.89 &0.88&0.89&0.89 & 0.87 &0.85 &0.87&0.87& 0.90&0.90&0.90&0.90
\\
    \midrule
   satellite & 0.80  &0.80&0.80&0.80&0.80&0.80&0.80&0.80&0.81&0.80&0.80&0.80
\\
    \midrule
   magic.gamma&0.78 &0.77&0.78&0.78 & 0.78&0.77&0.78&0.78& 0.78&0.78&0.78&0.78
\\
     \midrule
  SpamBase & 0.94&0.94&0.94&0.94 & 0.94&0.93&0.94&0.94&0.94&0.94&0.94&0.94
\\ 
    \midrule
     satimage-2& 0.97 &0.98&0.98&0.98& \textbf{0.93}&0.98&0.98&0.98& 0.98&0.99&0.99&0.99
\\
     \midrule
mammography& 0.90 &0.91&0.91&0.91 & 0.90&0.91&0.91&0.91& 0.90&0.91&0.91&0.91
\\ 
     \midrule
     vowels  & \textbf{0.77}  &0.85&0.87&0.87 &\textbf{0.69}&\textbf{0.77}&0.85&0.85& 0.85 &0.88&0.88&0.88
\\
    \midrule
   CIFAR10-1 &\textbf{0.69} &0.73&0.77&0.77& \textbf{0.66}& \textbf{0.71} & 0.76 &0.76 &\textbf{0.71}&0.74&0.77&0.77
\\  
     \midrule
     SVHN-1 & 0.80&0.82&0.84&0.84& \textbf{0.79}& 0.82& 0.84 & 0.84 & 0.80 & 0.83 & 0.84 & 0.84
  \\
     \midrule
    20news-1 & \textbf{0.64} &0.67&0.70&0.70&\textbf{0.56}&\textbf{0.59}&0.65&0.66&0.72&0.75&0.75&0.75
\\
     \midrule
      agnews-1& 0.94 &0.96&0.97&0.97&\textbf{0.88}&0.91&0.95&0.96&0.96&0.98&0.98&0.98
\\
    \midrule
     amazon  & \textbf{0.72 } &0.78&0.82&0.82&\textbf{0.66}&0.72&0.77&0.77&\textbf{0.76}&0.80&0.84&0.84
\\
    \midrule
   imdb &\textbf{ 0.75}&0.80&0.83&0.83&\textbf{0.69}&0.74 &0.79&0.80&\textbf{0.78}&0.82&0.85&0.85
  \\
    \midrule
    yelp&\textbf{0.82} &0.87&0.90&0.90&\textbf{0.74}&0.80&0.85&0.86&\textbf{0.85}&0.89&0.92&0.92\\
     \midrule
    vertebral & 0.71 &0.71&0.72&0.72&0.71&0.70&0.70&0.71&0.72&0.73&0.73&0.72
  \\
    \midrule
    fault & 0.65  &0.63&0.65&0.65&0.63&0.60&0.63&0.64&0.66&0.65&0.66&0.66
\\ 
    \bottomrule
    \end{tabular} %
    }    
\end{table}

Table \ref{tab:shallow_4} reports the mean of the AUC of shallow rAD over the 30 trials for different values of $a$.

\begin{table}\captionsetup{font=small}
\caption{AUC means of shallow rAD over 30 trials for different $a$. The significant changes in the AUC means are highlighted in bold.}
    \label{tab:shallow_4}
\centering
    \adjustbox{scale=0.75}{%
        \begin{tabular}{cccccccccc}
    \toprule
   \multirow{2}{*}{Dataset}  & \multicolumn{3}{c}{square\big/$a$ } &  \multicolumn{3}{c}{hinge\big/$a$ } & \multicolumn{3}{c}{m-Huber\big/$a$} \\
   \cmidrule{2-10}
        & 0.3 & 0.7& 0.9 &  0.3 & 0.7& 0.9 & 0.3 & 0.7& 0.9  \\
              \midrule
              thyroid &0.996&0.99&0.99&0.996&0.99&0.99&0.996&0.99&0.99 
\\
     \midrule
  pendigits &0.98&0.98&0.98&0.98&0.98&0.98&0.98&0.98&0.98 
\\ 
     \midrule
     Waveform &0.84&0.81&0.80&0.82&0.80&\textbf{0.77}&0.85&0.83&0.81
\\
    \midrule
   optdigits &0.997&0.995&0.99&0.996&0.995&0.99&0.997&0.996&0.99 
\\  
     \midrule
     Stamps &0.83&0.82&0.82&0.81&0.82&0.81&0.80&0.81&0.81 
\\
     \midrule
    mnist &0.97&0.96&0.96&0.96&0.96&0.96&0.97&0.96&0.96 
\\
     \midrule
     cardio &0.92&0.91&0.91&0.88&0.87&0.85&0.93&0.93&0.93 
\\
    \midrule    campaign&0.85&0.85&0.85&0.85&0.85&0.85&0.85&0.85&0.85
\\
    \midrule
  InternetAds  &0.79&\textbf{0.69}&\textbf{0.62}&0.87&0.85&\textbf{0.77}&0.83&\textbf{0.71}&\textbf{0.65}
\\
    \midrule
  landsat & 0.74&0.74&0.73& 0.74&0.74&0.73& 0.74&0.74&0.74
\\
     \midrule
    Cardiotocography & 0.87& 0.87& 0.87 &0.86&0.85&0.83&0.90&0.89&0.88
\\
    \midrule
   satellite &0.80&0.80&0.80&0.80&0.80&0.80&0.80&0.81&0.81
\\
    \midrule
   magic.gamma &0.78 &0.78&0.78&0.78&0.78&0.77&0.78&0.78&0.78
\\
     \midrule
  SpamBase &0.94&0.94&0.93 &0.94&0.94&0.93&0.94&0.94&0.94
\\ 
    \midrule
     satimage-2& 0.98& 0.99& 0.98& 0.98& 0.99& 0.98& 0.99& 0.99& 0.98
\\
     \midrule
mammography& 0.91 & 0.91& 0.91& 0.91& 0.91& 0.91& 0.91& 0.91& 0.91
\\ 
     \midrule
     vowels  & 0.87& 0.85& 0.83 & 0.85& 0.81& 0.78& 0.88& 0.87& 0.86
\\
    \midrule
   CIFAR10-1 &0.76&0.73&0.70&0.75&0.74&0.72&0.76&0.72&\textbf{0.69}
\\  
     \midrule
     SVHN-1 &0.84&0.83&0.82&0.83&0.83&0.82&0.84&0.83&0.81
  \\
     \midrule
    20news-1 &0.71&0.69&\textbf{0.65}&0.63&0.61&0.60&0.76&0.70&\textbf{0.66}
\\
     \midrule
      agnews-1& 0.98& 0.97& 0.96& 0.94& 0.94& 0.94& 0.98& 0.98& 0.97
\\
    \midrule
     amazon &0.81&0.80&0.77&0.77&0.75&0.76&0.83&0.81&0.79
\\
    \midrule
   imdb &0.82&0.80&0.78&0.78&0.77&0.75&0.84&0.81&0.79
  \\
    \midrule    yelp&90&0.88&0.86&0.84&0.83&0.82&0.91&0.89&0.87\\
     \midrule    vertebral&0.72&0.73&0.73&0.73&0.74&0.73&0.74&0.75&0.74
  \\
    \midrule
    fault &0.65&0.65&0.65&0.63&0.64&0.64&0.66&0.66&0.66
\\ 
    \bottomrule
    \end{tabular} %
    }    
\end{table}

\subsection{Additional experiments for deep rAD}
\begin{figure}[ht]
    \centering
    \includegraphics[width=0.9\textwidth]{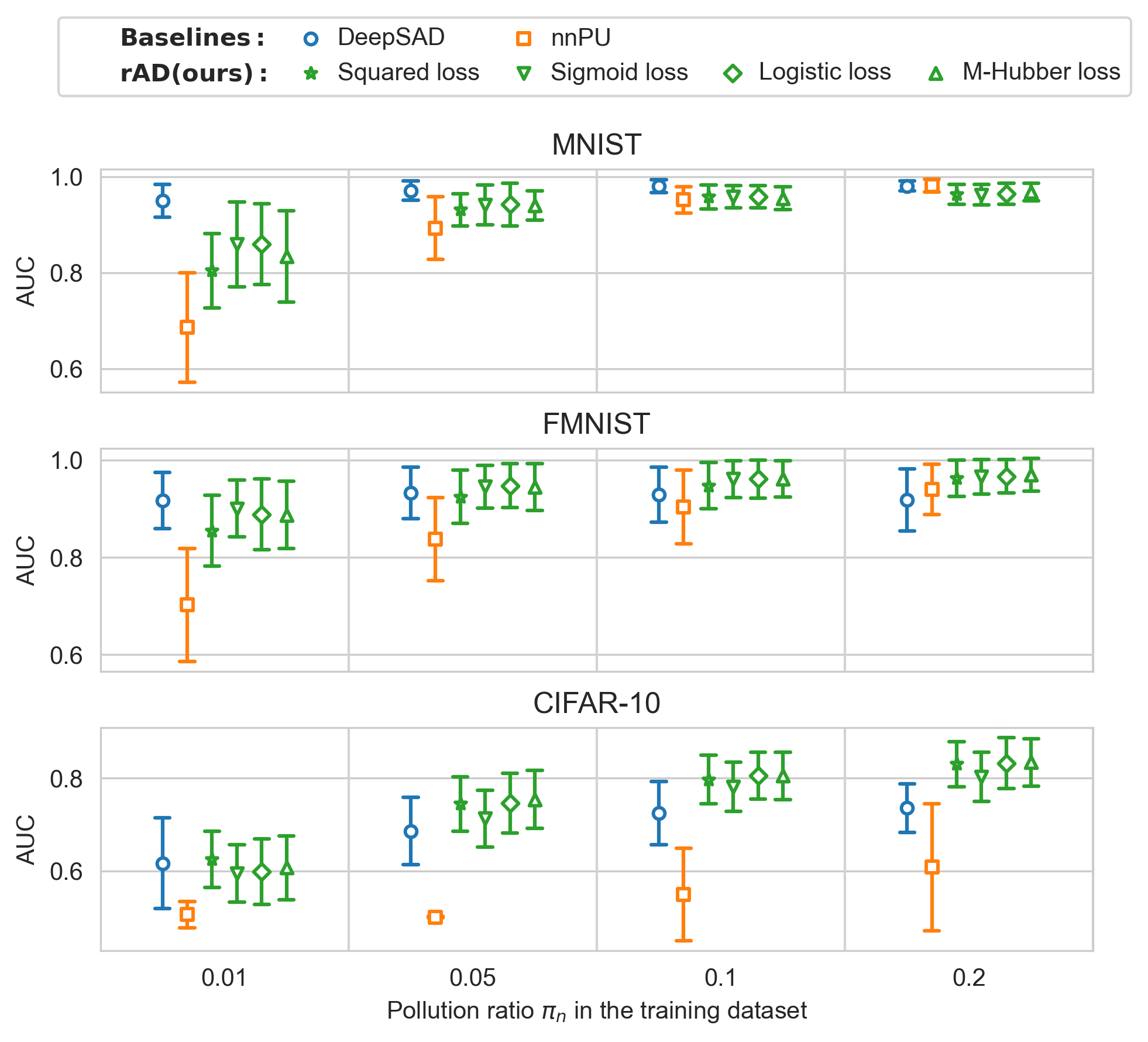}
    \caption{AUC mean and std over 20 trials with various $\pi_n$ and default $\gamma_l = 0.1$}
    \label{fig:results_gamma0.1}
\end{figure}
\textbf{Additional experiments for $\gamma_l\in\{0.1, 0.2\}$} \quad 
Figure \ref{fig:results_gamma0.1} and  Figure \ref{fig:results_gamma0.2} respectively report the mean and standard error (std) of the AUC over 20 trials on 3 benchmark datasets with increasing pollution ratio $\pi_n$ and $\gamma_l=0.1$ and  $\gamma_l=0.2$. 
\begin{figure}[ht]
    \centering
    \includegraphics[width=0.9\textwidth]{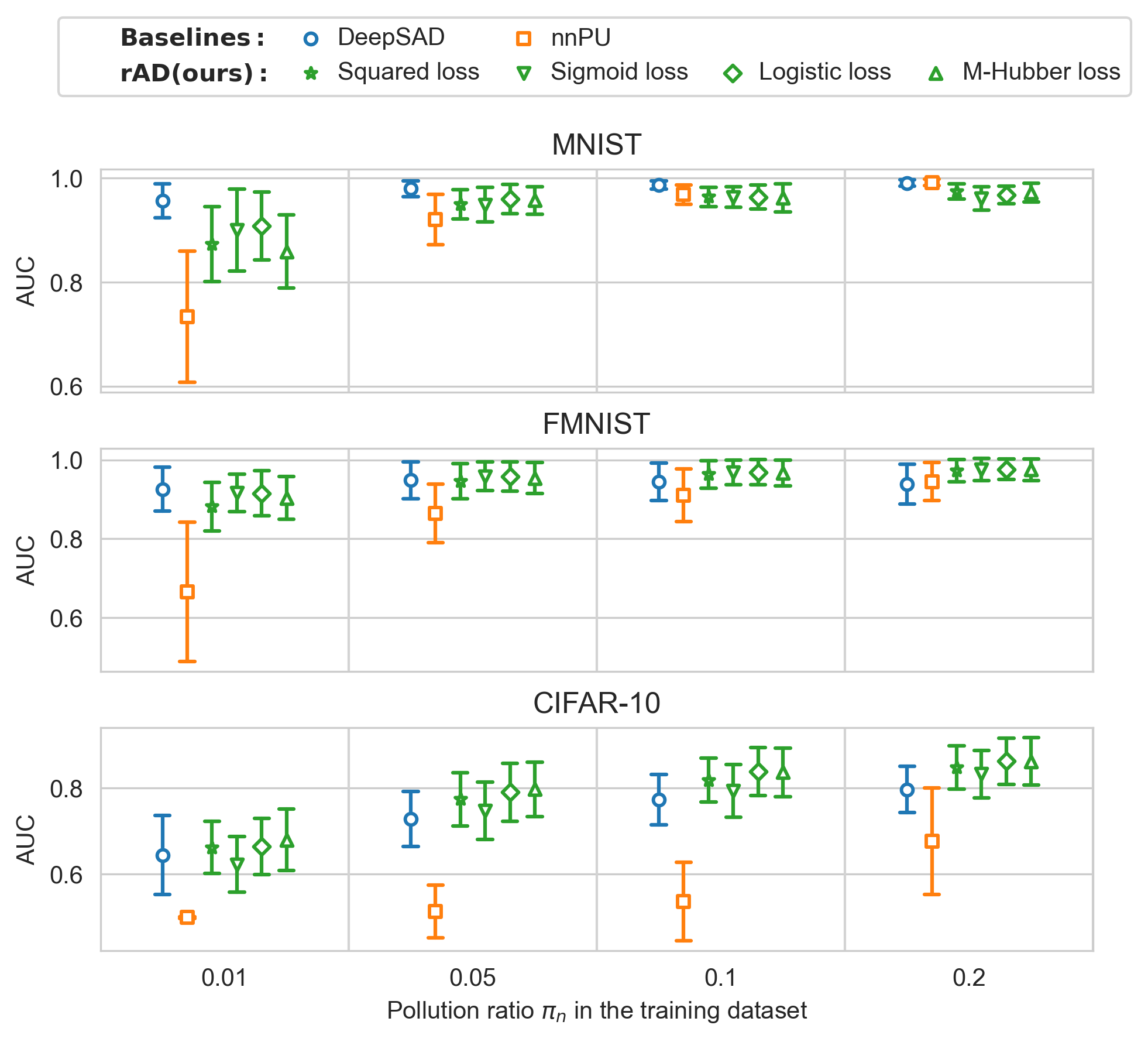}
    \caption{AUC mean and SE over 20 trials with various $\pi_n$ and default $\gamma_l = 0.2$}
    \label{fig:results_gamma0.2}
\end{figure}

\textbf{Impact of $\gamma_l$} \quad Figure \ref{fig:results_difgamma}  reports the AUC mean and std over 20 trials at various $\gamma_l$ for fixing $\pi_n = 0.1$.
 
\begin{figure}[ht]
    \centering
    \includegraphics[width=0.8\textwidth]{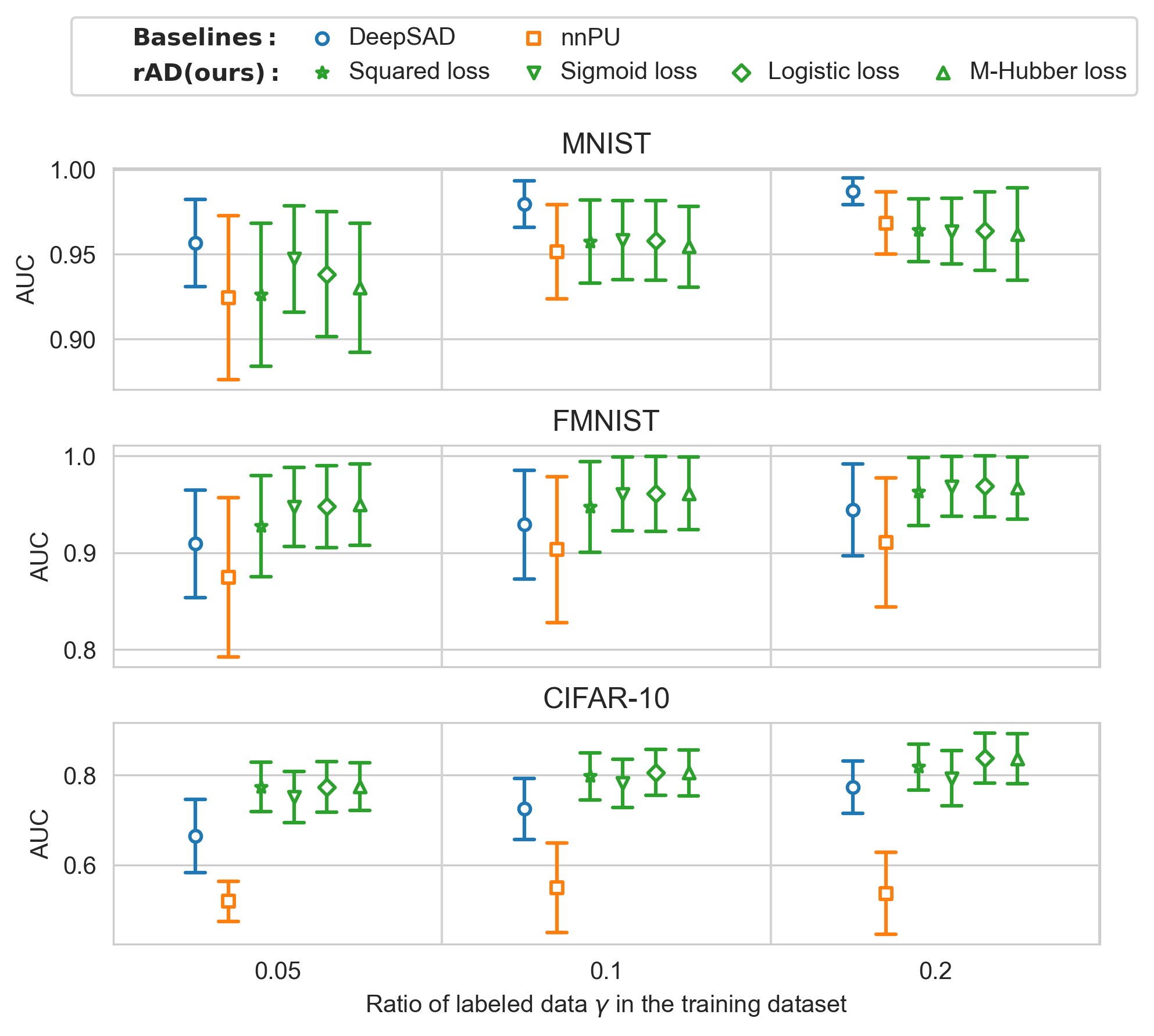}
    \caption{AUC mean and std over 20 trials at various $\gamma_l$ for fixing $\pi_n = 0.1$}
    \label{fig:results_difgamma}
\end{figure}

\textbf{Sensitivity analysis for $\pi_p^e$} \quad 
Table \ref{tab:deep_1} reports the mean and the standard error of the AUC of deep rAD over the 20 trials for different values of $\pi_p^e$. 

\begin{table}\captionsetup{font=small}
\caption{AUC means (and standard error) of deep rAD over 20 trials for different $\pi_p^e$. The significant changes in the AUC means are highlighted in bold.}
    \label{tab:deep_1}
\centering
    \adjustbox{scale=0.8}{%
        \begin{tabular}{ccccccc}
        \toprule
       Dataset & Loss & $\pi_p^e=1-\pi_n$ & $\pi_p^e=0. 9$ & $\pi_p^e=0.8$ & $\pi_p^e=0.7$ & $\pi_p^e=\pi_n$ \\ \midrule
       \multirow{4}{*}{\thead{MNIST \\($\pi_n=0.01$)}} 
       &square&\textbf{0.66}(0.04)&0.70(0.03)&0.72(0.02)&0.68(0.03)&\textbf{0.65}(0.03) \\
       \cmidrule{2-7}
       &sigmoid&\textbf{0.68}(0.03)&0.76(0.03)&0.76(0.03)&0.77(0.03)&0.77(0.03) \\
       \cmidrule{2-7}
       &logistic&\textbf{0.67}(0.03)&0.76(0.03)&0.80(0.03)&0.77(0.03)&0.77(0.03) \\
       \cmidrule{2-7}
       &m-Huber&\textbf{0.68}(0.03)&0.74(0.03)&0.71(0.03)&0.72(0.03)&0.73(0.03) \\
       \midrule
       \multirow{4}{*}{\thead{MNIST \\($\pi_n=0.05$)}} 
       &square&0.85(0.02)&0.87(0.01)&0.89(0.01)&0.89(0.01)& 0.86(0.01)\\
       \cmidrule{2-7}
       &sigmoid&0.88(0.01)&0.91(0.01)&0.91(0.01)&0.93(0.01)& 0.93(0.01)\\
       \cmidrule{2-7}
       &logistic&0.87(0.02)&0.89(0.01)&0.92(0.01)&0.92(0.01)&0.91(0.01) \\
       \cmidrule{2-7}
       &m-Huber&0.86(0.01)&0.88(0.01)&0.90(0.01)&0.90(0.01)& 0.87(0.01)\\
       \midrule
       \multirow{4}{*}{\thead{MNIST\\ ($\pi_n=0.1$)}} 
       &square&0.92(0.01)&0.92(0.01)&0.93(0.01)&0.93(0.01)&\textbf{ 0.89}(0.01)\\
       \cmidrule{2-7}
       &sigmoid&0.94(0.01)&0.94(0.01)&0.95(0.01)&0.95(0.01)&0.94(0.01) \\
       \cmidrule{2-7}
       &logistic&0.93(0.01)&0.93(0.01)&0.94(0.01)&0.94(0.01)&0.93(0.01) \\
       \cmidrule{2-7}
       &m-Huber&0.93(0.01)&0.93(0.01)&0.93(0.01)&0.94(0.01)& 0.92(0.01)\\
       \midrule\multirow{4}{*}{\thead{MNIST \\($\pi_n=0.2$)}}
       &square&0.95(0.01)&0.95(0.01)&0.95(0.01)&0.96(0.01)&0.94(0.01) \\
       \cmidrule{2-7}
       &sigmoid&0.95(0.01)&0.95(0.01)&0.95(0.01)&0.95(0.01)&0.95(0.01) \\
       \cmidrule{2-7}
       &logistic&0.96(0.01)&0.96(0.01)&0.96(0.01)&0.96(0.01)&0.96(0.01) \\
       \cmidrule{2-7}
       &m-Huber&0.95(0.01)&0.94(0.01)&0.95(0.01)&0.95(0.01)&0.94(0.01) \\
             \midrule
        \multirow{4}{*}{\thead{F-MNIST \\($\pi_n=0.01$)}} 
        &square&\textbf{0.76}(0.02)&0.80(0.01)&0.83(0.02)&0.84(0.02)&\textbf{0.78}(0.02) \\
       \cmidrule{2-7}
       &sigmoid&0.86(0.02)&0.87(0.02)&0.87(0.02)&0.88(0.02)& 0.88(0.02)\\
       \cmidrule{2-7}
       &logistic&0.85(0.02)&0.87(0.02)&0.87(0.02)&0.88(0.02)& 0.87(0.02)\\
       \cmidrule{2-7}
       &m-Huber&\textbf{0.82}(0.02)&0.88(0.02)&0.87(0.02)&0.87(0.02)& \textbf{0.83}(0.02)\\
       \midrule
       \multirow{4}{*}{\thead{F-MNIST \\($\pi_n=0.05$)}}
       &square&\textbf{0.84}(0.01)&0.86(0.01)&0.89(0.01)&0.91(0.01)&0.91(0.01) \\
       \cmidrule{2-7}
       &sigmoid&0.93(0.01)&0.93(0.01)&0.93(0.01)&0.94(0.01)&0.95(0.01) \\
       \cmidrule{2-7}
       &logistic&0.91(0.01)&0.92(0.01)&0.93(0.01)&0.93(0.01)& 0.95(0.01)\\
       \cmidrule{2-7}
       &m-Huber&0.92(0.01)&0.94(0.01)&0.93(0.01)&0.93(0.01)&0.93(0.01) \\
       \midrule
       \multirow{4}{*}{\thead{F-MNIST\\ ($\pi_n=0.1$)}} 
       &square&\textbf{0.88}(0.01)&\textbf{0.88}(0.01)&0.93(0.01)&0.94(0.01)&0.94(0.01) \\
       \cmidrule{2-7}
       &sigmoid&0.94(0.01)&0.94(0.01)&0.95(0.01)&0.95(0.01)&0.96(0.01) \\
       \cmidrule{2-7}
       &logistic&0.94(0.01)&0.94(0.01)&0.95(0.01)&0.95(0.01)& 0.96(0.01)\\
       \cmidrule{2-7}
       &m-Huber&0.95(0.01)&0.95(0.01)&0.95(0.01)&0.95(0.01)&0.95(0.01) \\
       \midrule
       \multirow{4}{*}{\thead{F-MNIST \\($\pi_n=0.2$)}} 
       &square&0.94(0.01)&0.92(0.01)&0.94(0.01)&0.95(0.01)&0.96(0.01) \\
       \cmidrule{2-7}
       &sigmoid&0.96(0.01)&0.95(0.01)&0.96(0.01)&0.96(0.01)&0.96(0.01) \\
       \cmidrule{2-7}
       &logistic&0.95(0.01)&0.94(0.01)&0.95(0.01)&0.96(0.01)& 0.97(0.01)\\
       \cmidrule{2-7}
       &m-Huber&0.96(0.01)&0.94(0.01)&0.96(0.01)&0.96(0.01)&0.96(0.01) \\
       \midrule
       \multirow{4}{*}{\thead{CIFAR-10 \\($\pi_n=0.01$)}} 
       &square&0.60(0.01)&0.60(0.01)&0.59(0.01)&0.60(0.01)&0.59(0.01) \\
       \cmidrule{2-7}
       &sigmoid&0.58(0.01)&0.58(0.02)&0.58(0.02)&0.57(0.02)&0.55(0.02) \\
       \cmidrule{2-7}
       &logistic&0.60(0.02)&0.58(0.02)&0.57(0.02)&0.56(0.02)&\textbf{0.53}(0.02) \\
       \cmidrule{2-7}
       &m-Huber&0.61(0.02)&\textbf{0.55}(0.02)&\textbf{0.55}(0.02)&\textbf{0.55}(0.02)& 0.60(0.02)\\
       \midrule
       \multirow{4}{*}{\thead{CIFAR-10 \\($\pi_n=0.05$)}} 
       &square&0.73(0.01)&0.72(0.01)&0.73(0.01)&0.72(0.01)&0.73(0.01) \\
       \cmidrule{2-7}
       &sigmoid&0.66(0.02)&0.68(0.01)&0.69(0.01)&0.67(0.01)&0.69(0.02) \\
       \cmidrule{2-7}
       &logistic&0.71(0.01)&0.71(0.02)&0.71(0.01)&0.70(0.01)& 0.69(0.01)\\
       \cmidrule{2-7}
       &m-Huber&0.69(0.01)&0.70(0.01)&0.71(0.01)&0.72(0.01)& 0.71(0.01)\\
       \midrule
       \multirow{4}{*}{\thead{CIFAR-10\\ ($\pi_n=0.1$)}} 
       &square&0.77(0.01)&0.77(0.01)&0.77(0.01)&0.78(0.01)& 0.77(0.01)\\
       \cmidrule{2-7}
       &sigmoid&0.75(0.01)&0.75(0.01)&0.75(0.01)&0.76(0.01)&0.73(0.01) \\
       \cmidrule{2-7}
       &logistic&0.77(0.01)&0.77(0.01)&0.77(0.01)&0.77(0.01)& 0.76(0.01)\\
       \cmidrule{2-7}
       &m-Huber&0.77(0.01)&0.77(0.01)&0.77(0.01)&0.78(0.01)& 0.77(0.01)\\
       \midrule
       \multirow{4}{*}{\thead{CIFAR-10 \\($\pi_n=0.2$)}} 
       &square&0.80(0.01)&0.80(0.01)&0.80(0.01)&0.80(0.01)& 0.80(0.01)\\
       \cmidrule{2-7}
       &sigmoid&0.77(0.01)&0.74(0.01)&0.77(0.01)&0.77(0.01)& 0.77(0.01)\\
       \cmidrule{2-7}
       &logistic&0.79(0.01)&0.78(0.01)&0.79(0.01)&0.80(0.01)&0.79(0.01) \\
       \cmidrule{2-7}
       &m-Huber&0.79(0.01)&0.79(0.01)&0.79(0.01)&0.80(0.01)& 0.80(0.01)\\
       \bottomrule
   \end{tabular} %
    }    
\end{table}

\textbf{Sensitivity analysis for $a$}  \quad Fixing $\pi_p^e=0.8$, Figure \ref{fig:results_a_0.01}--\ref{fig:results_a_0.2} show AUC mean and std of deep rAD with additional values of $a\in\{0.5, 0.9\}$ ($a=0.1$ is the default setting) on the  datasets with $\gamma_l = 0.05$ and $\pi_n=\in \{0.01, 0.05, 0.2\}$ 

\begin{figure}
    \centering
    \includegraphics[width=0.6\textwidth]{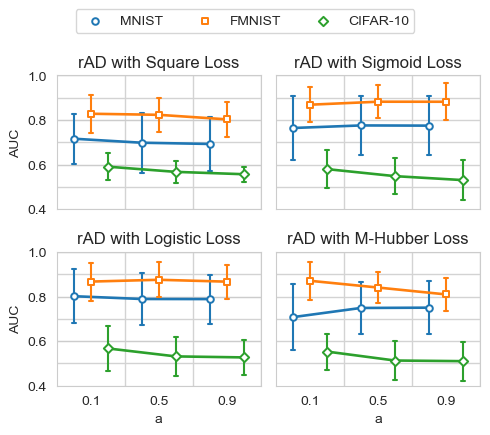}
    \caption{AUC mean and std over 20 trials at various $a$ for the datasets with  $\gamma_l = 0.05$ and $\pi_n = 0.01$.}
    \label{fig:results_a_0.01}
\end{figure}

\begin{figure}
    \centering
    \includegraphics[width=0.6\textwidth]{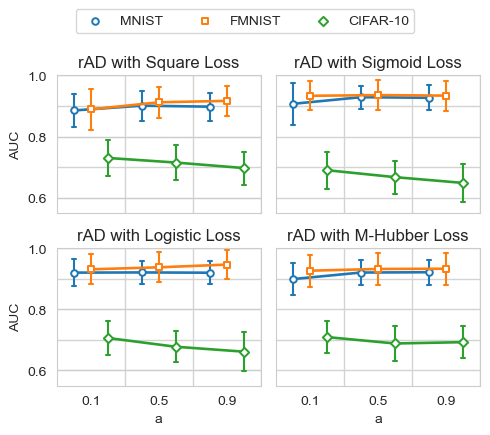}
    \caption{AUC mean and std over 20 trials at various $a$ for the datasets with $\gamma_l = 0.05$ and $\pi_n = 0.05$.}
    \label{fig:results_a_0.05}
\end{figure}

\begin{figure}
    \centering
    \includegraphics[width=0.6\textwidth]{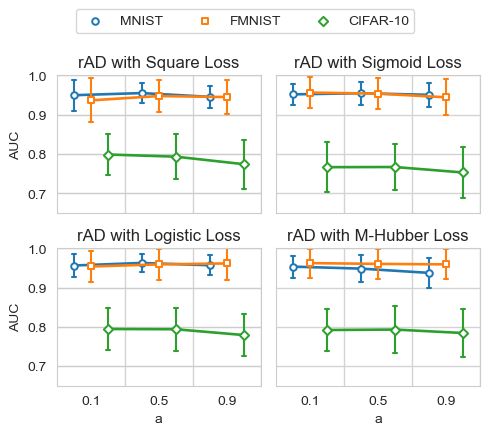}
    \caption{AUC mean and std over 20 trials at various $a$ for the datasets with $\gamma_l = 0.05$ and $\pi_n = 0.2$.}
    \label{fig:results_a_0.2}
\end{figure}
\textbf{ROC curves } \quad 
Figure \ref{fig:roc} shows representative ROC curves obtained by a trial of running the methods (with default settings) on the datasets with $\gamma = 0.05$ and $\pi_n = 0.1$. 

\begin{figure}
    \centering
    \includegraphics[width=\textwidth]{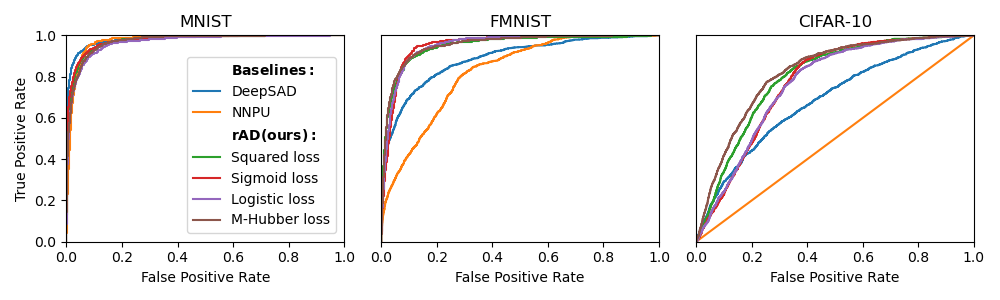}
    \caption{Representative ROC curves for different datasets with $\gamma = 0.05$ and $\pi_n = 0.1$.}
    \label{fig:roc}
\end{figure}
\fi
\end{document}